# Rank-Based Learning and Local Model Based Evolutionary Algorithm for High-Dimensional Expensive Multi-Objective Problems

Guodong Chen, Jiu Jimmy Jiao, Xiaoming Xue and Zhongzheng Wang

*Abstract*— Surrogate-assisted evolutionary algorithms have been widely developed to solve complex and computationally expensive multi-objective optimization problems in recent years. However, when dealing with high-dimensional optimization problems in decision space, the performance of these surrogate-assisted multi-objective evolutionary algorithms deteriorate drastically. In this work, a novel Classifier-assisted rank-based learning and Local Model based multi-objective Evolutionary Algorithm (CLMEA) is proposed for high-dimensional expensive multi-objective optimization problems. CLMEA makes full use of the uncertainty of solutions in the decision space and objective space to explore the uncertain but informative space towards high-dimensional problems. Specifically, the offspring in different ranks uses rank-based learning strategy to generate more promising and informative candidates for real function evaluations. In order to reduce the search space of high-dimensional problems and maintain the diversity of solutions, the most uncertain sample point from the non-dominated solutions measured by the crowding distance is selected as the center to conduct local search. The experimental results of benchmark problems and a real-world application on geothermal reservoir heat extraction optimization demonstrate superior performance of CLMEA compared with the state-of-the-art surrogate-assisted multi-objective evolutionary algorithms. The source code for this work is available at https://github.com/JellyChen7/CLMEA.

## I. Introduction

MULTI-OBJECTIVE optimization has attracted extensive research concerns in recent years since many real-world optimization problems contain several conflicting objectives [1]. Multi-objective evolutionary algorithms (MOEAs), inspired by the evolution mechanism of organisms, provide powerful black-box optimizers to search for the optimal Pareto solutions [2-4], and have been widely applied to deal with many complex practical optimization problems. Most existing frameworks of the MOEAs can mainly be classified into three types [5]: dominance-based MOEAs, indicator-based MOEAs, and decomposition-based MOEAs. Dominance-based MOEAs use Pareto dominance to differentiate and sort solutions, such as NSGA-II [6] and MOPSO [7]. Indicator-based MOEAs utilize the performance indicators, e.g., hypervolume (HV) [8] and inverted generational distance (IGD) [9], as the infill criteria. Decomposition-based MOEAs, such as MOEA/D [10], decompose an MOP into a set of sub-problems and optimize them simultaneously. Nevertheless, many real-world engineering problems involve time- or resource-intensive evaluations, e.g., computational fluid dynamics simulations [11], thermal-hydraulic-mechanical coupling simulations [12], and physical and chemical experiments [13]. Evolutionary algorithms become computationally prohibitive due to the large number of function evaluations (FEs) required before convergence when tackling expensive multi-objective optimization problems (EMOPs).

Machine learning techniques have made remarkable progress in solving EMOPs [1, 14, 15], which can be mainly classified into three categories from the purpose of the models: surrogate-assisted methods, estimation of Pareto front distributions, and learning-assisted offspring generation. Table I summarizes representative machine learning models employed in surrogate-assisted or learning-assisted MOEAs and corresponding experimental settings and test suites. Surrogate-assisted evolutionary algorithms (SAEAs) have shown their effectiveness in reducing the number of FEs during the optimization process, and have been extensively developed in addressing EMOPs in the past decade [14, 16]. According to the output of surrogates, SAEAs can be roughly divided into approximation-based surrogates and classification-based surrogates. Approximation-based surrogates, including polynomial response surface [17], Gaussian process (also known as Kriging) [18-20], radial basis function network (RBF) [21-24], artificial neural network [25] and support vector regression (SVR) [26-28], build computationally efficient mathematical models to approximate optimization landscapes of interest. It is not trivial to decide the surrogate type without any prior information since it is problem-dependent [29]. Wang *et al.* [29] combined polynomial response surface, Kriging, and RBF as the ensemble surrogate to solve single-objective

This work is supported by General Fund of Natural Science Foundation of Guangdong Province No. 2019A151511021, and grants from the HKU Seed Fund for Basic Research. (Corresponding author: Jiu Jimmy Jiao)

Guodong Chen and Jiu Jimmy Jiao are with the Faculty of Sciences, The University of Hong Kong, Hong Kong SAR, China (e-mail: u3008598@connect.hku.hk; jjiao@hku.hk)

Xiaoming Xue is with the Department of Computer Science, City University of Hong Kong, Hong Kong SAR, China (e-mail: xminghsueh@gmail.com).

Zhongzheng Wang is with the College of Engineering, Peking University, China (e-mail: wangzzgr@163.com).

expensive optimization problems. MOEA/D-EGO [19] decomposed an MOP into several sub-problems, and employed kriging model to maximize the expected improvement metric for each sub-problem. K-RVEA [30] used Kriging to approximate objective functions, and generated reference vectors to guide the evolutionary search. NSGAIII-EHVI [31] combined the framework of NSGA-III and Kriging surrogate model, infilling with expected HV improvement criterion, to solve many-objective problems. KTA2 [32] introduced an adaptive infill criterion to identify the most important requirement on convergence, diversity, or uncertainty to deal with EMOPs. RVMM [33] employed an adaptive model management strategy assisted by two sets of reference vectors to calculate an amplified upper confidence bound. MASTO [34] used an adaptive technique to dynamically establish the most promising RBF and Kriging surrogates, and employed multiple infill criteria to solve MOPs. EMMOEA [35] constructed surrogate model with Kriging model for each objective and developed a new performance indicator that balances the diversity and convergence of the algorithm to select promising solutions for real FEs. Nevertheless, most Kriging-assisted evolutionary algorithms is computationally intensive and can mainly be applied to problems with less than 15 decision variables [36-38]. END-ARMOEA [25] was proposed using dropout neural network to replace Gaussian process as surrogate towards high-dimensional many-objective optimization problems.

Classification-based surrogates, such as support vector classifier [27, 39], gradient boosting classifier [40], artificial neural network [41, 42], and k-nearest neighbor (KNN) [43], learn the dominance relationship between the sample points to select promising solutions. If the dominance relationships between the Pareto solutions and the offspring candidates are known, fitness values would be less useful, and environment selection can be performed efficiently. CSEA [41] employed feedforward neural network (FNN) to predict the dominance relationship between the selected reference solutions and candidate solutions. PARETO-SVM [44] used SVM to predict the dominance relationship and tightly characterize the current Pareto set and the dominated region. In contrast, MOEA/D-SVM [45] constructed a classifier according to the value of the scalarization function for each sub-problem, and pre-select new generated solutions for real FEs. CPS-MOEA [43] built a KNN model to classify non-dominated solutions as good solutions, and pre-screened candidate solutions to perform real FEs. MCEA/D [39] employed multiple local classifiers as scalarization function, obtained from decomposition for high-dimensional MOPs with dimensions varying from 50 to 150. Nevertheless, the pre-screening ability of classifiers is relatively poor, since the prediction is less informative in comparison with approximation methods [39].

Estimation of Pareto front distributions is able to mitigate the adverse risk of sparse area in Pareto front and improve the diversity of the Pareto solutions [46, 47]. Commonly used methods to approximate the Pareto set are Bayesian network [48], regression decision tree [49], generative adversarial network [50], and manifold learning [46]. Tian *et al.* [51] developed a Pareto front estimation method based on local models to guide the search direction of evolutionary algorithms. Li and Kwong [46] applied a principal curve algorithm to obtain an approximation of the Pareto solutions manifold in solving EMOPs. Li *et al.* [47] introduced a Pareto front model-based local search method called SMEA-PF. This method built a Pareto front model with current optimal non-dominated solutions, and selected some sparse points to preform local surrogate assisted search.

TABLE I
SUMMARY OF MACHINE LEARNING METHODS INVOLVED IN MOEAs.

| Algorithm | Machine learning model | | | Experimental settings | | | Test suits |
|---|---|---|---|---|---|---|---|
| | Method | Usage | Purpose | D | M | $FEs_{max}$ | |
| MOEA/D-EGO [15] | Kriging | Approx. | Objective functions | 2, 6, 8 | 2, 3 | 200, 300 | ZDT, LZF, DTLZ |
| K-RVEA [25] | Kriging | Approx. | Objective functions | 10 | 3, 4, 6, 8, 10 | 300 | DTLZ |
| NSGAIII-EHVI [26] | Kriging | Approx. | Objective functions | 10 | 3, 5, 8, 10 | 100 | DTLZ, IDTLZ, WFG |
| KTA2 [27] | Kriging | Approx. | Objective functions | 9, 10, 11 | 3, 4, 6, 8, 10 | 300 | DTLZ, WFG |
| RVMM [28] | Kriging | Approx. | Objective functions | 5, 6, 10, 11, 13, 20 | 2, 3, 5, 9, 10 | 300, 400, 500, 600, 800 | DTLZ, DPF, WFG, MaF, DDMOP |
| MASTO [29] | Kriging, RBF | Approx. | Objective functions | 5, 10, 20, 30 | 2, 3 | 300 | DTLZ, ZDT |
| EMMOEA [30] | Kriging | Approx. | Objective functions | 10, 11 | 3, 5, 10 | 300 | DTLZ, WFG, MaF |
| SMEA-PF [42] | RBF | Approx. | Objective functions | 8-30 | 2, 3 | 200, 300 | DTLZ, ZDT, Reactive power optimization |
| ESF-KVEA [16] | Kriging | Approx. | Objective functions | 10, 20, 30 | 2, 3 | 11D+120 | DTLZ, WFG, UF, MaF |
| CSEA [36] | FNN | Class. | Pareto dominance | 9, 10, 11, 20, 30 | 2–4, 6, 8–10 | 300, 600, 900 | DTLZ |
| PARETO-SVM [39] | SVM | Class. | Pareto dominance | 10, 30 | 2 | 1E5 | IHR, ZDT |
| MOEA/D-SVM [40] | SVM | Class. | Scalarization function | 10, 30 | 2, 3 | 1E5, 3E5 | DTLZ, UF, ZDT |
| CPS-MOEA [38] | KNN | Cluster | Pareto dominance | 30 | 2, 3 | 2000, 4E4, 1E5 | DTLZ, ZDT |
| HeE-MOEA [31] | SVM, RBF, PCA | Approx. | Objective functions | 10, 20, 40, 80 | 3 | 11D+119 | DTLZ, WFG |
| EDN-ARMOEA [20] | DNN | Approx. | Objective functions | 20, 40, 60, 100 | 3, 5, 10, 20 | 11D+119 | DTLZ, WFG |
| MCEA/D [34] | SVM | Class. | Scalarization function | 50, 100, 150 | 3, 7, 11 | 300 | DTLZ, WFG |
| GMOEA [45] | GAN | Generator | Offspring Generation | 30, 50, 100, 200 | 2, 3 | 5000, 1E4, 1.5E4, 3E4 | IMF |
| ALMOEA [48] | MLP | Generator | Offspring Generation | 1000, 3000, 5000, 10000 | 2, 3 | 1E4, 15300 | LMF, LSMOP |
| CLMEA | RBF, PNN | Approx., Class. | Objective functions, Scalarization function | 30, 50, 100, 160, 200 | 2, 3 | 300 | DTLZ, ZDT, Geothermal optimization |

Learning-assisted offspring generation learns and explores the landscape information of problems in the design space to accelerate evolutionary search and converge into promising area efficiently [52]. Liu *et al.* [2] systematically overviewed recent advances of learnable MOEAs and summarized the attractive new directions from a unique perspective. He *et al.* [50] developed GMOEA driven by generative adversarial networks to generate high-quality solutions in high-dimensional decision space. Liu *et al.* [53] proposed an accelerated evolutionary search algorithm ALMOEA, where a multilayer perceptron is adopted to learn a gradient-descent-like direction vector for each solution to reproduce promising solutions. Zhan *et al.* [54] developed a learning-aided evolutionary optimization framework that integrates evolution knowledge learned by neural network from the evolution process to accelerate the convergence. Tian *et al.* [55] proposed an operator selection method based on reinforcement learning for evolutionary multi-objective optimization. Zhen *et al.* [56] also used reinforcement learning to select surrogate-assisted sampling strategy in solving expensive single-objective optimization problems. Wang *et al.* [57] developed deep reinforcement learning for generalizable oil reservoir well-control optimization problems. The aforementioned reinforcement learning algorithms are valuable explorations in learnable optimization framework. However, they highly rely on the evolutionary operators or surrogate-assisted sampling, which cannot ideally evolve and generate the solutions independently. How to design a more intelligent and learnable optimization framework still requires more efforts.

To further accelerate the convergence of optimizers on high-dimensional EMOPs, a novel classifier-assisted rank-based learning and local model-based multi-objective evolutionary algorithm namely CLMEA is proposed. The proposed algorithm consists of three parts: classifier-assisted rank-based learning pre-screening, HV-based non-dominated search, and local search in the sparse objective space. The main contributions of this paper can be summarized as follows:

1) To the best of our knowledge, this is the first time to introduce probabilistic neural network model as a classifier to rank the candidate solutions, and guide the further offspring generation. Considering the classifier is not enough to generate strong active offspring evolution pressure, rank-based learning strategy is developed to generate more promising and informative candidates, and the most uncertain first-ranking solution in the decision space is selected to conduct real FEs. In contrast to conventional classification-based algorithms, CLMEA presents a classifier-assisted rank-based learning strategy that generates promising offspring solutions under the guidance of classifier rather than generating offspring and then pre-screening promising offspring solutions with a classifier. Rank-based learning strategy can enhance the generation of elite solutions and the selection of real FEs used the uncertainty of solutions in decision space.

2) The ways of selecting center or range of the local search can impact the performance of the algorithm significantly. In order to reduce the search space of high-dimensional problems and maintain the diversity of solutions, local surrogates centered at the sparse point of current front selected by the crowding distance are constructed to infill solutions at the sparse non-dominated front. The most uncertain non-dominated solutions are selected to conduct real FEs, thus improving the diversity of solutions.

3) CLMEA is compared with five state-of-the-art surrogate-assisted MOEAs on high-dimensional MOPs. Experimental results on DTLZ and ZDT problems with dimensions varying from 30 to 200 and a real-world application on heat extraction optimization of fractured geothermal reservoir demonstrate that the proposed algorithm shows better convergence and diversity performance than other state-of-the-art MOEAs.

The rest of this article is organized as follows. Section II presents preliminaries of this article. The proposed classifier and local model based evolutionary algorithm is described in Section III. Experimental results and analysis are illustrated in Section IV. Conclusions and discussions are provided in Section V.

## II. PRELIMINARIES

This section first presents a brief problem definition, then introduces probabilistic neural network and radial basis function surrogates. Finally, the motivation of this work is briefly described.

### A. Problem Definition

Without loss of generality, an unconstrained MOP can be mathematically modeled as

$$\min_{\boldsymbol{x}} \boldsymbol{F}(\boldsymbol{x}) = (f_1(\boldsymbol{x}), f_2(\boldsymbol{x}), ..., f_m(\boldsymbol{x})) \quad (1)$$
$$s.t. \ \boldsymbol{x} \in \Omega$$

where $\boldsymbol{x} = (x_1, x_2, ..., x_d)$ is the decision vector with $d$ variables in a feasible region $\Omega$, and $\boldsymbol{F}$ is a set of $m$ objective

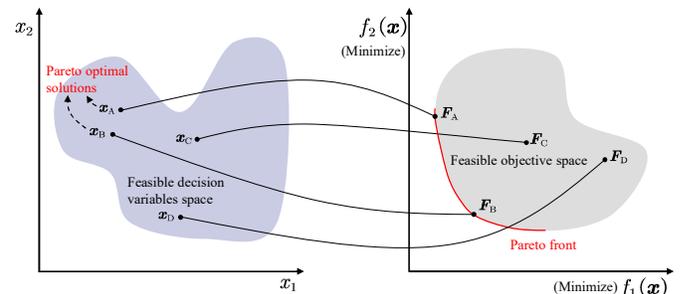

Fig. 1. Schematic diagram of MOPs and related concepts.

functions. Fig. 1 presents the schematic diagram of MOPs and related concepts, with the following definitions involved:

**Definition 1** (Pareto dominance): For any two solutions $x_1, x_2 \in \Omega$, $x_1$ dominates $x_2$ iff $f_i(x_1) \leq f_i(x_2)$ for $\forall i \in \{1, 2, ..., m\}$ and $f_j(x_1) < f_j(x_2)$ for $\exists j \in \{1, 2, ..., m\}$, referred to as $x_1 \succ x_2$. For instance, in Fig. 1, $x_B$ dominates $x_C$, while $x_A$ and $x_B$ are mutually non-dominated.

**Definition 2** (Pareto optimal solutions and Pareto front): Any solution $x \in \Omega$ is said to be Pareto optimal solution iff there is no other solution dominating it. Pareto optimal solutions in a set is known as Pareto set. The projection of the Pareto set into the objective space is known as Pareto front.

When the number of objectives $m$ is larger than 3, the problems are known as many objective problems (MaOPs) [58]. The performance of MOEAs deteriorate dramatically when solving MaOPs since the non-dominated solutions increase exponentially with the number of objectives, causing dominance-based MOEAs fail to identify the solutions [41]. Note that expensive optimization problems denote that the objective calculation involves time-consuming numerical simulations or resource-intensive experiments. In the context of expensive optimization problems, dimensions above 30 are considered high-dimensional expensive problems. Most existing SAEAs were developed to solve EMOPs with dimensions less than 30.

### B. Probabilistic Neural Network

Probabilistic neural network (PNN) [59], developed from Bayesian decision strategy, is a feed-forward neural network that is able to map the input pattern into any number of categories. PNN is composed of four layers, an input layer, a pattern layer, a summation layer and an output layer. The input layer receives the input vector with $d$ variables. In the case of Gaussian kernel, the pattern layer computes the probability density estimation from the input data to the kernel center as:

$$\varphi_{ij}(x) = \frac{1}{2\pi^{d/2}\sigma^d} \exp\left[-\frac{(x - C_{ij})(x - C_{ij})}{2\sigma^2}\right] \quad (2)$$

where $\sigma$ is the smoothing parameter, $C_{ij}$ is the $j^{th}$ training pattern from the category $i$. Then the summation layer calculates the maximum likelihood of $x$ being classified into the $i^{th}$ category by:

$$p_i(x) = \frac{1}{N_i} \sum_{j=1}^{N_i} \varphi_{ij}(x) \quad (3)$$

where $N_i$ is the number of training vectors of the $i^{th}$ category. Furthermore, the output layer chooses the maximum $p_i(x)$ as the prediction of the classifier.

### C. Radial Basis Function

RBF, a weighted sum of basis functions, has been widely employed as surrogates to approximate the nonlinear landscape of the objective functions in the past decades. Given a set of training sample points $\{[x_1, y_1], ...[x_i, y_i], ..., [x_k, y_k]\}$, the surrogate model is built as:

$$\hat{f}(x) = \sum_{i=1}^{k} \omega_i \psi(\|x - c_i\|) \quad (4)$$

where $\{\omega_i\}_{i=1}^{k}$ is the weight coefficient, $\psi(\cdot)$ is the basis function, and $c_i$ is the center vector of the basis function. Note that $c_i$ is set as $x_i$ and the Gaussian basis function is employed in this work. The weight coefficient can be determined as:

$$\omega = \Phi^{-1} y \quad (5)$$

where $\Phi = [\psi(\|x_i - x_j\|)]_{k \times k}$ is the kernel matrix.

### D. Motivations

The main motivation of this work comes from the need of the real-world simulation-based MOPs, i.e., heat extraction optimization of geothermal reservoir, hydrocarbon production optimization under geological uncertainty, and hydrological design optimization. Only limited works have been done towards high-dimensional EMOPs [25, 39]. Since high-dimensional problems require more computing resources to converge, this contradicts the expensive property of the problem. The deterioration of surrogate quality in high-dimensional space also exacerbates the contradiction. As discussed in Section I, the existing surrogate-assisted MOEAs still have some room to better explore and exploit the search space for improvement in solving EMOPs, especially for high-dimensional problems. Most MOEAs involving machine learning methods, as summarized in TABLE I, target on problems with dimension no larger than 100D. MCEA/D extends the dimension to 150, with promising optimization performance on benchmark functions [39]. GMOEA [50] and ALMOEA [53] attempted to solve problems with 30-200D and 1000-10000D, respectively, while the computational budgets are larger.

This work aims to extend the dimension of EMOPs to 200 with very little computation budget (FEs<=300) by applying stronger evolution pressure of the offspring and exploring uncertain but promising region and making full use of the uncertainty of solutions in the decision space and objective space to explore the uncertain but informative space. In contrast to conventional classification-based algorithms, we present a classifier-assisted rank-based learning strategy that generates promising offspring solutions under the guidance of classifier rather than generating offspring and then pre-screening promising offspring solutions with a classifier. Rank-based learning strategy can enhance the generation of elite solutions and the selection of real FEs used the uncertainty of solutions in decision space. To further accelerate the exploitation of the population in MOEAs and to guarantee the diversity of the achieved nondominated solutions, we develop HV-based non-dominated search to achieve faster convergence in early period. To reduce the search space of high-dimensional problems and maintain the diversity of solutions, local surrogates centered at the sparse point of current front are constructed to infill solutions at the sparse non-dominated front. The most uncertain non-dominated solutions in the objective space are selected for real FEs, thus improving the diversity of solutions.

## III. CLASSIFIER-ASSISTED RANK-BASED LEARNING AND LOCAL MODEL BASED EVOLUTIONARY ALGORITHM

In this section, the framework of the proposed CLMEA is introduced in detail. Classifier-assisted rank-based learning pre-screening, HV-based non-dominated search, and local search in the sparse objective space are subsequently described.

### A. Framework

The pseudo-code and framework diagram of the proposed CLMEA are presented in Algorithm 1 and Fig. 2, respectively. Initially, the sample points are generated using Latin hypercube sampling (LHS) from the decision space, and the real function evaluations are performed. Then add all evaluated sample points into the archive $A = \{x_i, y_i\}$. Classifier-assisted rank-based learning pre-screening is first developed to generate more promising and informative candidates, which uses non-dominated rank and distance information, for real function evaluations. Subsequently, a HV-based non-dominated search is performed to efficiently speed up the convergence. After searching the non-dominated solutions of the surrogate model, the candidates with higher HV improvement are selected for real FEs. Furthermore, in order to keep the diversity of the

**Algorithm 1** Proposed CLMEA

**Input:** The maximum number of function evaluations $maxFEs$; the number of initial sample points $N$; the population size $NP$; the number of infill solutions $n$
**Output:** Non-dominated solutions
    1) Generate $N$ initial sample points $\{x_1,...,x_N\}$ with LHS
    2) Evaluate real function values of $\{x_1,...,x_N\}$
    3) $FEs \leftarrow N$
    4) Initialize $A \leftarrow \{(x_1, y_1),...,(x_N, y_N)\}$
    5) For i = 1:m
    6)    $\hat{x}_{\text{extreme}} \leftarrow \arg\min(f_i(x))$
    7)    Conduct real function evaluation on $\hat{x}_{\text{extreme}}$
    8)    $FEs \leftarrow FEs + 1$; $A \leftarrow A \cup (\hat{x}_{\text{extreme}}, y_{\text{extreme}})$
    9) End for
    10) While $FEs < maxFEs$
    11)    $\{\hat{x}_{c1},...,\hat{x}_{cn}\} \leftarrow$ Perform classifier-assisted rank-based learning pre-screening    // **Algorithm 2**
    12)    Conduct real function evaluations on $\{\hat{x}_{c1},...,\hat{x}_{cn}\}$
    13)    $FEs \leftarrow FEs + n$; $A \leftarrow A \cup \{(\hat{x}_{c1}, y_{c1}),...,(\hat{x}_{cn}, y_{cn})\}$
    14)    $\{\hat{x}_{h1},...,\hat{x}_{hn}\} \leftarrow$ Execute HV-based non-dominated search    // **Algorithm 3**
    15)    Evaluate the function values of $\{\hat{x}_{h1},...,\hat{x}_{hn}\}$
    16)    $FEs \leftarrow FEs + n$; $A \leftarrow A \cup \{(\hat{x}_{h1}, y_{h1}),...,(\hat{x}_{hn}, y_{hn})\}$
    17)    $\{\hat{x}_{l1},...,\hat{x}_{ln}\} \leftarrow$ Conduct local search in the sparse objective space    // **Algorithm 4**
    18)    Evaluate the function values of $\{\hat{x}_{l1},...,\hat{x}_{ln}\}$
    19)    $FEs \leftarrow FEs + n$; $A \leftarrow A \cup \{(\hat{x}_{l1}, y_{l1}),...,(\hat{x}_{ln}, y_{ln})\}$
    20) End while

solutions, the most uncertain sample points from the non-dominated solutions measured by crowding distance is selected as the guided parent for the generation of promising candidate solutions. Then the neighbor points in the objective space are adopted as the parents for further offspring generation. The non-dominated solutions are pre-screened by the surrogate, and the sparse solutions of the current non-dominated solutions are selected for real FEs to further infill in the sparse region of the current non-dominated front. The optimization process continues until the maximum number of FEs is reached.

### B. Classifier-Assisted Rank-Based Learning Pre-screening

Classifier-assisted rank-based learning pre-screening is mainly used to enhance the exploitation of Pareto front, and also promote the exploration of sparse and promising areas. The detailed pseudo-code and algorithm process of classifier-assisted rank-based learning pre-screening strategy is presented in Algorithm 2 and Fig. 3, respectively. Specifically, non-dominated sorting is executed for the solutions in the archive $A$, and top $NP$ solutions are selected as the population $P \leftarrow \{x_1,...,x_{NP}\}$ based on the non-dominated sorting and crowding distance metrics. Solutions to the first level are labeled 1, solutions to the second level are labeled 2, until all the solutions in $P$ are ranked $\{l_1,...,l_{NP}\}$. After that, a PNN is constructed as classifier to predict the offspring. Rank-based learning operator is then developed to generate more promising and informative candidates:

$$v_i = x_{r_1} + Mu \times (x_{r_2} - x_{r_3}) \qquad (6)$$

where $x_{r_1}$ and $x_{r_2}$ are individuals randomly selected from the first level, $x_{r_3}$ is an individual randomly selected from the first or the second level. Subsequently, crossover operation and polynomial mutation are conducted to generate offspring $u$. PNN is then employed to rank the offspring candidates. The loop continues until the proportion of offspring belonging to the first level is higher than 0.9. After the loop ends, the uncertainty of each offspring candidate is calculated based on the Euclidean distance to the nearest evaluated sample points:

$$g(u_i) = \min_{x \in A} \{dis(u_i,x)\} \qquad (7)$$

where $dis(u_i,x)$ is the Euclidean distance between the offspring $u_i$ and evaluated sample points $x$ in $A$. The most uncertain offspring is selected as follows:

$$\hat{x}_c = \arg\max_{u_i \in P} g(u_i) \qquad (8)$$

where $\hat{x}_c$ is the selected offspring candidates to be evaluated.

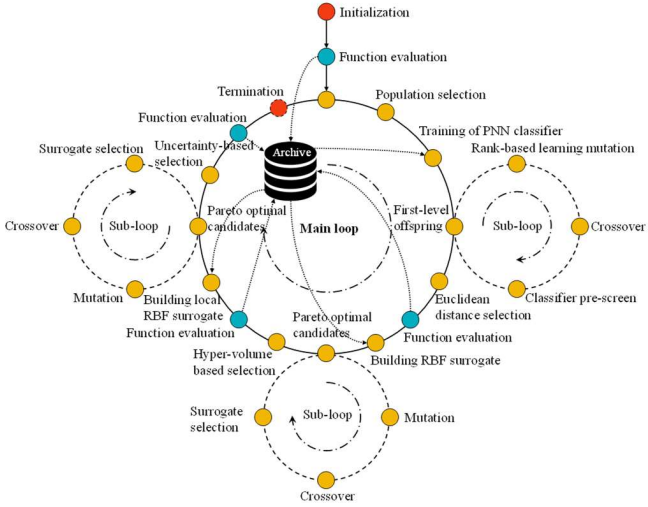

Fig. 2. Framework diagram of the proposed CLMEA consisting of three sub-loops. In the main loop, the candidate sample points are selected to perform real function evaluations, while in the sub-loop, the candidate solutions are selected based on the prediction of the surrogates.

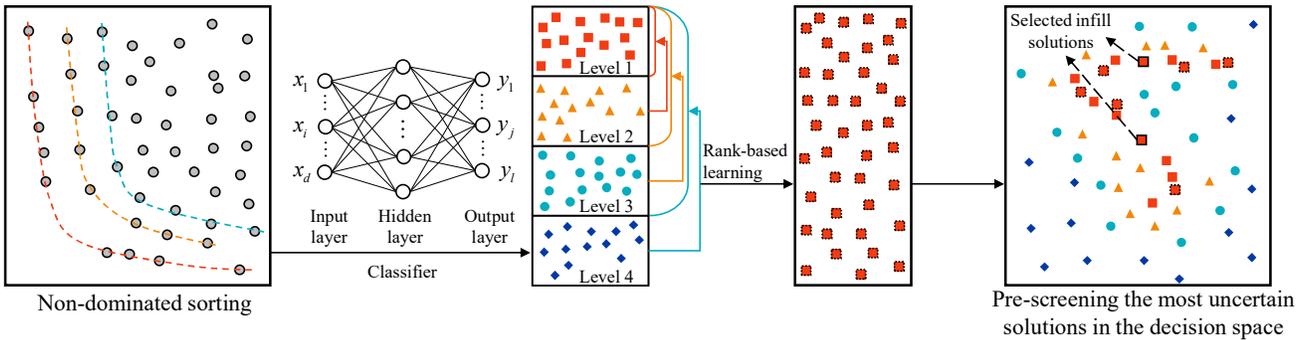

Fig. 3. The process of classifier-assisted rank-based learning pre-screening

**Algorithm 2** Classifier-Assisted Rank-Based Learning Pre-screening
---
**Input:** The population size $NP$, archive $A$, the number of infill solutions $n$
**Output:** The selected offspring candidates $\{\hat{x}_{c1},...,\hat{x}_{cn}\}$.
    1) Execute non-dominated sorting for the solutions in the archive $A$
    2) Select top $NP$ solutions using non-dominated sorting and crowding distance as the population $P \leftarrow \{x_1,...,x_{NP}\}$
    3) Rank the solutions $P$ into several levels $\{l_1,...,l_{NP}\}$ according to the Pareto sort
    4) Train a PNN classifier using the ranking data $\{(x_1,l_1),...,(x_{NP},l_{NP})\}$
    5) While $sum(l_i=1)/NP < 0.9$
    6)    Generate offspring $P \leftarrow \{x_1,...,x_{NP}\}$ with rank-based learning operator in eq. 6, crossover operator and polynomial mutation
    7)    Predict the level of offspring $\{\hat{l}_1,...,\hat{l}_{NP}\}$ candidates using PNN
    8) End while
    9) Calculate the uncertainty of each offspring candidate with eq. 7
    10) Select the top $n$ candidates with maximum uncertainty $\{\hat{x}_{c1},...,\hat{x}_{cn}\}$

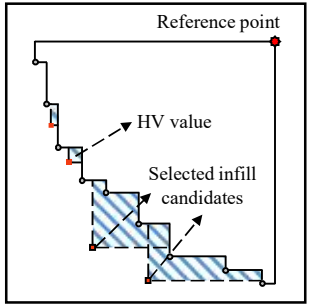

Fig. 4. Schematic diagram for the infill criterion of HV-based non-dominated search

**Algorithm 3** Hypervolume-Based Non-Dominated Search
---
**Input:** The maximum number of evolving generations $max\_gen_1$, archive $A$, the number of infill solutions $n$
**Output:** The selected offspring candidates $\{\hat{x}_{h1},...,\hat{x}_{hn}\}$.
    1) Select top $NP$ solutions in $A$ as the population $P \leftarrow \{x_1,...,x_{NP}\}$ based on non-dominated sorting and crowding distance
    2) Construct RBF surrogates for all objectives with all sample points
    3) For i = 1:$max\_gen_1$
    4)    $Q \leftarrow$ Reproduction using binary and polynomial mutation operator and crossover operator
    5)    Predict the objective function values with RBF surrogates
    6)    $P \leftarrow$ Conduct environmental selection
    7) End for
    8) Calculate the HV improvement of each candidate in $P$
    9) Choose top $n$ candidates with maximum uncertainty $\{\hat{x}_{h1},...,\hat{x}_{hn}\}$

### C. Hypervolume-Based Non-Dominated Search

HV-based non-dominated search mainly targets at accelerating the convergence of the optimization process, and is also able to enhance the diversity of the solutions. The pseudo-code of HV-based non-dominated search is shown in Algorithm 3. Concretely, RBF surrogates are constructed for all the objectives. Non-dominated sorting DE is employed to search for the Pareto front of the surrogates. Polynomial mutation operator and binary crossover operator are used to generate the offspring, and surrogates are employed to predict the objective value for the environmental selection of new populations $P$. The evolution of populations ends after $max\_gen_1$ generations. After that, HV criterion is adopted to select the most promising solutions. Schematic diagram for the infill criterion of HV-based non-dominated search is shown in Fig. 4. The blue shaded area indicates the HV improvement of candidate solutions predicted by the surrogates. The HV

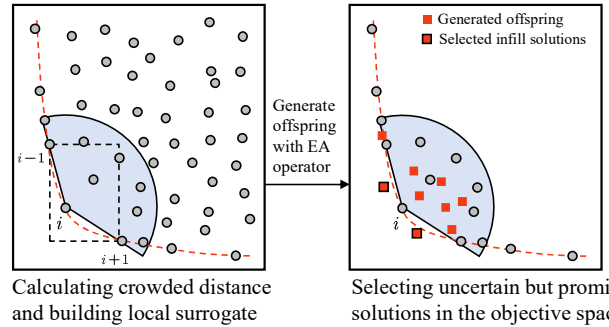

Fig. 5. Schematic diagram for the infill criterion of local search in the sparse objective space

improvement of each candidate in $P$ can be calculated using the existing Pareto front as follows:

$$\hat{x}_h = \underset{\hat{x}_i \in P}{arg\ max}\ [HV(x_{A1} \cup \hat{x}_i) - HV(x_{A1})] \quad (9)$$

where $\hat{x}_h$ is the selected offspring to be evaluated for HV-based non-dominated search strategy, $HV$ denotes the HV calculation, and $x_{A1}$ is the first rank solutions in archive $A$.

### D. Local Search in the Sparse Objective Space

Local search in the sparse objective space aims to identify and refine the sparse region in the current non-dominated front. The sparse point in the non-dominated front can be identified and further refined by generating promising solutions nearby with the help of the local surrogates. Algorithm 4 presents the procedure of the proposed infill criterion. A simple schematic diagram is illustrated in Fig. 5 to facilitate the understanding of the local search in the sparse objective space. Specifically, the sparsest points at the current Pareto front are selected based on non-dominated sorting and crowding distance of each solution in current Pareto front. Subsequently, local surrogates around the sparse points are constructed. The sparse points are adopted as the guided parent to further infill in the sparse region. After evolving the populations guided by local surrogates, the first rank candidate solutions are pre-screened using the prediction of surrogates. After that, the uncertainty of each candidate solutions in the objective space is calculated:

$$l(\hat{x}_i) = \underset{x \in A}{min}\ \{dis(f(\hat{x}_i), f(x))\} \quad (10)$$

where $\hat{x}_i$ is the $ith$ candidate solution in the population, and $f$ is the objective function vector of a solution. The most uncertain offspring in objective space is selected as follows:

TABLE II
AVERAGE IGD VALUES OF CLMEA AND ITS VARIANTS ON BI-OBJECTIVE DTLZ PROBLEMS.

| Problems | D | CLMEA-s1 | CLMEA-s2 | CLMEA-s3 | CLMEA |
|---|---|---|---|---|---|
| DTLZ1 | 30 | 4.5152e+02 | 3.3713e+02 | 5.7399e+02 | **2.8640e+02** |
|  | 50 | 8.9079e+02 | 5.2922e+02 | 1.1085e+03 | **4.5536e+02** |
|  | 100 | 2.2942e+03 | 9.5045e+02 | 2.5769e+03 | **9.3303e+02** |
|  | 200 | 4.8879e+03 | 2.2131e+03 | 5.3505e+03 | **2.1620e+03** |
| DTLZ2 | 30 | 1.4864e+00 | 5.6201e-01 | 9.3186e-01 | **2.9013e-01** |
|  | 50 | 2.3844e+00 | 1.0661e+00 | 1.6370e+00 | **8.9291e-01** |
|  | 100 | 5.2137e+00 | **1.2047e+00** | 4.6590e+00 | 1.7226e+00 |
|  | 200 | 1.0235e+01 | **2.2710e+00** | 9.2077e+00 | 3.1795e+00 |
| DTLZ3 | 30 | 1.1301e+03 | 7.0434e+02 | 1.5121e+03 | **6.7609e+02** |
|  | 50 | 2.1888e+03 | 1.2320e+03 | 2.9071e+03 | **1.1955e+03** |
|  | 100 | 6.3368e+03 | 2.9594e+03 | 6.7058e+03 | **2.6658e+03** |
|  | 200 | 1.3936e+04 | 5.8530e+03 | 1.4050e+04 | **5.7911e+03** |
| DTLZ4 | 30 | 1.6392e+00 | 1.0586e+00 | 1.2786e+00 | **1.0051e+00** |
|  | 50 | 2.6903e+00 | 1.3095e+00 | 1.7593e+00 | **1.2718e+00** |
|  | 100 | 5.1767e+00 | 2.1619e+00 | 3.9507e+00 | **1.8990e+00** |
|  | 200 | 1.0699e+01 | **3.1819e+00** | 8.0876e+00 | 3.2875e+00 |
| DTLZ5 | 30 | 1.3782e+00 | 3.6742e-01 | 8.5387e-01 | **1.8035e-01** |
|  | 50 | 2.5841e+00 | 6.9148e-01 | 1.8346e+00 | **2.1626e-01** |
|  | 100 | 5.0783e+00 | **1.1169e+00** | 4.7836e+00 | 1.4172e+00 |
|  | 200 | 9.9787e+00 | **2.0794e+00** | 9.6019e+00 | 3.2023e+00 |
| DTLZ6 | 30 | 1.5125e+01 | **8.1515e+00** | 2.3219e+01 | 9.3408e+00 |
|  | 50 | 2.8674e+01 | **1.7556e+01** | 4.0057e+01 | 1.9042e+01 |
|  | 100 | 6.8984e+01 | 5.4892e+01 | 8.3479e+01 | **4.4918e+01** |
|  | 200 | 1.4387e+02 | 1.1897e+02 | 1.6834e+02 | **1.0259e+02** |
| DTLZ7 | 30 | 5.6019e+00 | 1.0769e+00 | 6.7245e+00 | **8.0399e-01** |
|  | 50 | 7.7316e+00 | 3.6422e+00 | 8.1116e+00 | **1.4570e+00** |
|  | 100 | 9.6778e+00 | 7.3396e+00 | 9.8122e+00 | **3.5307e+00** |
|  | 200 | 1.0292e+01 | 8.2758e+00 | 1.0505e+01 | **6.3128e+00** |
| +/ − / ≈ |  | 0/ 28 / 0 | 7/ 21 / 0 | 0/ 28 / 0 | NA |

$$\hat{x}_l = \underset{\hat{x}_i \in P}{arg\ max}\ g(\hat{x}_i) \quad (11)$$

This sampling strategy can further infill in the uncertain region of the non-dominated front and maintain the diversity of the final optimal solutions.

## IV. EXPERIMENTAL STUDIES

In this section, the performance of the proposed CLMEA is first investigated regarding the efficacy of classifier-assisted rank-based learning pre-screening, HV-based non-dominated search, and local search in sparse object space. CLMEA is also compared with five state-of-the-art algorithms (CPS-MOEA [43], K-RVEA [47], CSEA [41], END-ARMOEA [25], and MCEA/D [39]) on DTLZ and ZDT benchmark suites (detailed characteristics of the benchmark problems in Table S-I in the supplementary material). The source code of CLMEA is publicly available in MATLAB to help readers reproduce the results [1]. Besides, all the experiments are conducted on evolutionary multi-objective optimization platform PlatEMO[2] [60]. In addition, the computational complexity of the algorithms is analyzed and compared. Finally, a real-world heat extraction optimization of geothermal reservoir is also employed to further test the performance of the proposed CLMEA.

---

**Algorithm 4** Local Search in the Sparse Objective Space

**Input:** The maximum number of evolution of generations $max\_gen_2$, the number of infill solutions $n$, the population size $NP$
**Output:** The selected offspring candidates $\{\hat{x}_{l1},...,\hat{x}_{ln}\}$.
  1) Conduct non-dominated sorting and calculate the crowding distance of each solution in current Pareto front
  2) Select top $n$ solutions except the endpoints of each objective.
  3) For i = 1: $n$
  4)    Choose $NP$ nearest points to the i[th] sparse point in the objective space
  5)    Construct local RBF surrogates for all objectives
  6)    For i = 1: $max\_gen_2$
  7)      $Q \leftarrow$ Reproduce the i[th] sparse point using binary and polynomial mutation operator and crossover operator
  8)      Predict the objective function values with RBF surrogates
  9)      $P \leftarrow$ Conduct environmental selection
  10)   End for
  11)   Perform non-dominated sorting with the current Pareto front and $P$
  12)   If $\exists \hat{x} \in P$ in the first sort
  13)      Calculate the uncertainty of each candidate solutions
  14)      Choose the most uncertain solution in objective space as $\hat{x}_{li}$
  15) End for

---

[1] https://github.com/JellyChen7/CLMEA

[2] https://github.com/BIMK/PlatEMO

## A. Parameter Settings

Seven DTLZ (DTLZ 1-7) and five ZDT (ZDT 1-4, 6) benchmark suites are employed in this work. Since this work targets high-dimensional multi-objective expensive problems, the number of objective functions and dimension of variables are set to $M=\{2,3\}$ and $D=\{30,50,100,200\}$, respectively. For all algorithms, the initial number of samples is set to 100 when the dimension is less than 100, and to 200 when the dimension is greater than or equal to 100. The termination criterion is the predefined FEs. In the experiments, the maximum number of FEs is set to 300 for all benchmarks. The performance of the algorithms is evaluated by IGD metric on the benchmark suites, while by HV metric on the real-world application. Wilcoxon signed-rank test is used with a significance probability $\alpha=0.05$ to compare the algorithms. The inner parameter settings of the compared algorithms keep unchanged except for the initial sampling number.

TABLE III
AVERAGE IGD RESULTS OF THE 20 INDEPENDENT RUNS ON BI-OBJECTIVE DTLZ PROBLEMS.

| Problems | D | CPS-MOEA | K-RVEA | CSEA | EDN-ARMOEA | MCEA/D | CLMEA |
|---|---|---|---|---|---|---|---|
| DTLZ1 | 30 | 6.1231e+02 | 5.9574e+02 | 5.3458e+02 | 7.6109e+02 | 3.7268e+02 | **3.0310e+02** |
|  | 50 | 1.1710e+03 | 1.2859e+03 | 1.1088e+03 | 1.4066e+03 | 6.8636e+02 | **5.3412e+02** |
|  | 100 | 2.7599e+03 | 3.0384e+03 | 2.9263e+03 | 3.1878e+03 | 1.9418e+03 | **1.0740e+03** |
|  | 200 | 5.9030e+03 | 6.7851e+03 | 6.3461e+03 | 6.7667e+03 | 4.0038e+03 | **1.9906e+03** |
| DTLZ2 | 30 | 1.2068e+00 | 1.2689e+00 | 8.4305e-01 | 1.3224e+00 | 4.3816e-01 | **9.7035e-02** |
|  | 50 | 1.8985e+00 | 2.9998e+00 | 1.7832e+00 | 2.8303e+00 | 6.2911e-01 | **1.3942e-01** |
|  | 100 | 5.7976e+00 | 6.4073e+00 | 5.4123e+00 | 6.4253e+00 | 1.9302e+00 | **3.8858e-01** |
|  | 200 | 1.1930e+01 | 6.3902e+00 | 1.2360e+01 | 1.3862e+01 | 3.9686e+00 | **1.1857e+00** |
| DTLZ3 | 30 | 1.6564e+03 | 1.6672e+03 | 1.3159e+03 | 2.1064e+03 | 7.2601e+02 | **7.0538e+02** |
|  | 50 | 3.1099e+03 | 3.5867e+03 | 2.7884e+03 | 3.8988e+03 | 1.5979e+03 | **1.1711e+03** |
|  | 100 | 7.4213e+03 | 8.4621e+03 | 7.9648e+03 | 8.7190e+03 | 4.0560e+03 | **2.6277e+03** |
|  | 200 | 1.5879e+04 | 1.8646e+04 | 1.7299e+04 | 1.8581e+04 | 8.9629e+03 | **5.2284e+03** |
| DTLZ4 | 30 | 1.5812e+00 | 1.6870e+00 | 9.3992e-01 | 1.4471e+00 | 8.0266e-01 | **6.7771e-01** |
|  | 50 | 2.2624e+00 | 3.2670e+00 | 2.0010e+00 | 2.9380e+00 | 9.0809e-01 | **7.3285e-01** |
|  | 100 | 5.7711e+00 | 6.6171e+00 | 5.6006e+00 | 6.5906e+00 | 1.9391e+00 | **8.8932e-01** |
|  | 200 | 1.1474e+01 | 1.4158e+01 | 1.2496e+01 | 1.4075e+01 | 3.6328e+00 | **9.5823e-01** |
| DTLZ5 | 30 | 1.2256e+00 | 1.2618e+00 | 8.5359e-01 | 1.3895e+00 | 4.0083e-01 | **1.0114e-01** |
|  | 50 | 1.9899e+00 | 2.9974e+00 | 1.7169e+00 | 2.8174e+00 | 6.3098e-01 | **1.4158e-01** |
|  | 100 | 5.5055e+00 | 6.4208e+00 | 5.4699e+00 | 6.3250e+00 | 2.0507e+00 | **4.6926e-01** |
|  | 200 | 1.2023e+01 | 1.3698e+01 | 1.2286e+01 | 1.3931e+01 | 4.1888e+00 | **1.3569e+00** |
| DTLZ6 | 30 | 1.9018e+01 | 2.1726e+01 | 2.0974e+01 | 2.2788e+01 | 1.2599e+01 | **1.1261e+01** |
|  | 50 | 3.4604e+01 | 3.9658e+01 | 3.8241e+01 | 4.0564e+01 | 2.4848e+01 | **2.0184e+01** |
|  | 100 | 7.6077e+01 | 8.4358e+01 | 8.3904e+01 | 8.5151e+01 | 5.9271e+01 | **4.6223e+01** |
|  | 200 | 1.5674e+02 | 1.7463e+02 | 1.7304e+02 | 1.7454e+02 | 1.1701e+02 | **9.5831e+01** |
| DTLZ7 | 30 | 5.8422e+00 | **6.7819e-02** | 3.1578e+00 | 2.0159e+00 | 5.2623e+00 | 4.7586e-01 |
|  | 50 | 6.2642e+00 | **1.9824e-01** | 3.9454e+00 | 3.0068e+00 | 6.2424e+00 | 8.9416e-01 |
|  | 100 | 6.9154e+00 | **1.6445e+00** | 5.5607e+00 | 4.8434e+00 | 6.8649e+00 | 1.8606e+00 |
|  | 200 | 7.2734e+00 | 7.1638e+00 | 6.3074e+00 | 6.1556e+00 | 7.2630e+00 | **5.0233e+00** |
| +/−/≈ |  | 0/ 28 / 0 | 3/ 25 / 0 | 0/ 28 / 0 | 0/ 28 / 0 | 0/ 28 / 0 | NA |

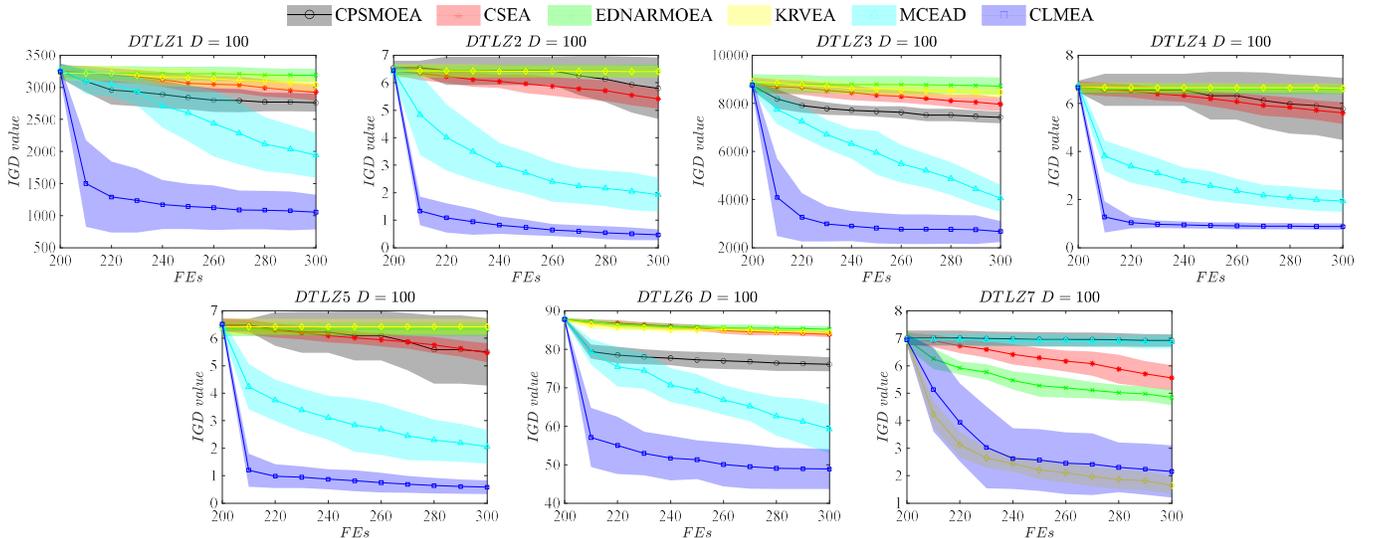

Fig. 6. Convergence curves of the six compared algorithms on 2-objective 100D DTLZ problems

For the parameter of CLMEA, the population size is set to 50. Offsprings are generated by DE mutation, polynomial mutation and binary crossover. The maximum number of evolutionary generations in the HV-based non-dominated search strategy is set to 50. The number of points for building local surrogate is set to 100 for problems with variables less than 100, and to 200 with variables greater than or equal to 100. The number of infill solutions is set to 1 for each infill strategy. The infill number can be larger if the decision maker wishes to make use of parallel computing power. The maximum number of evolution of generations for local search in the sparse objective space is set to 10. Besides, all the experiments are performed on MATLAB R2021a.

### B. Performance metrics

The HV indicator and inverted generational distance (IGD) are the most frequently applied performance metrics to quantitatively assess the convergence and diversity of the final population. IGD calculates the average distance from each Pareto optimal solution to the nearest solution. Suppose $P^*$ is a set of evenly distributed reference solutions on the Pareto front, and $Q$ is a set of non-dominated solutions. The IGD is mathematically defined as:

$$IGD(P^*, Q) = \frac{\sum_{p \in P^*} \text{dis}(p, Q)}{|P^*|} \quad (12)$$

where $\text{dis}(p, Q)$ denotes the minimum Euclidean distance between $p$ and solutions in $Q$ provided by MOEAs, and $|P^*|$ denotes the number of reference points. The smaller IGD value indicates better approximation to the true Pareto front. The calculation of IGD value requires the distribution of Pareto front. Nevertheless, for many real-world scientific and engineering problems, the explicit information of Pareto front is not available.

HV indicator [61] is able to measure the dominated volume of the populations in the objective space without the prior knowledge of Pareto front distribution. The HV improvement can be adopted as the performance metric [62] and the infill criterion [27, 31] of SAEAs for MOPs. Suppose that reference vector $z_r = [z_{r1}, ..., z_{rm}]$ is a worst point dominated by all the Pareto optimal objective vectors, the HV can be calculated as:

$$HV(Q, z) = L([z, x_1] \cup ... \cup [z, x_k]) \quad (13)$$

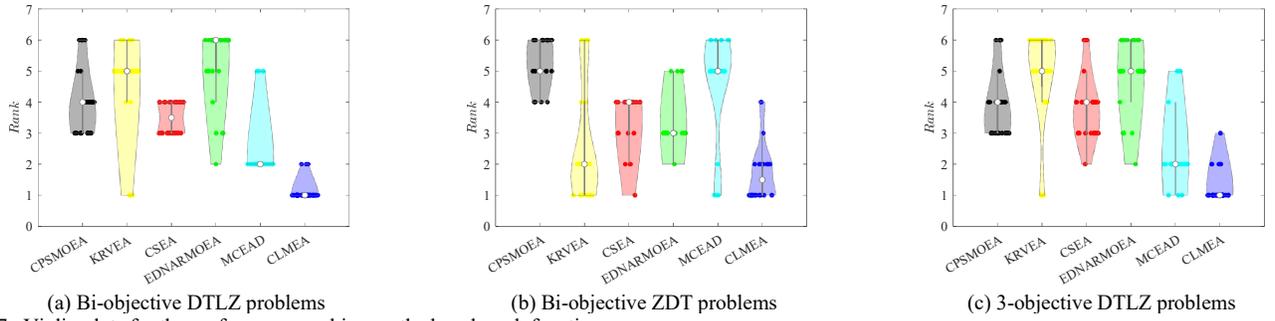

(a) Bi-objective DTLZ problems    (b) Bi-objective ZDT problems    (c) 3-objective DTLZ problems

Fig. 7. Violin plots for the performance ranking on the benchmark functions

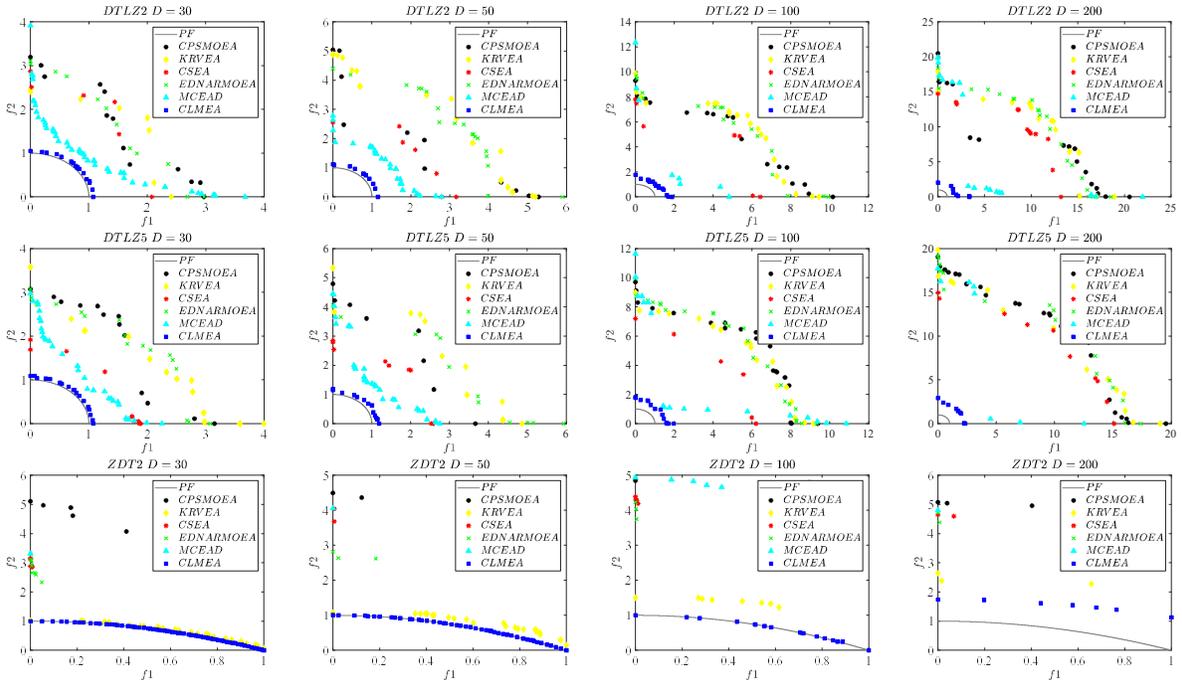

Fig. 8. Non-dominated front by CLMEA and compared algorithms on bi-objective benchmark problems

where $L$ is the usual Lebesgue measure, and $\{x_1,...,x_k\}$ are the solutions in $Q$ provided by MOEAs. The larger HV indicates better approximation to the true Pareto front. The drawback of HV indicator is the computational cost especially on MaOPs [63].

### C. Effectiveness of the Three Sampling Strategies

CLMEA is first compared with its variants CLMEA-s1, CLMEA-s2, and CLMEA-s3, which use classifier-assisted rank-based learning pre-screening, HV-based non-dominated search, and local search in the sparse objective space, respectively. Average IGD values of CLMEA and its variants on bi-objective DTLZ problems, with dimensions 30, 50, 100 and 200, are shown in Table II. The results on bi-objective ZDT and 3-objective DTLZ problems are listed in Table S-II and S-III respectively. All the results of benchmark functions are replicated 20 independent runs for statistical analysis. '+', '−', and '≈' indicate that the result is statistically significantly better, worse and comparable to CLMEA, respectively. From Table II, CLMEA outperforms CLMEA-s1 and CLMEA-s3 on all benchmark functions, while worse than CLMEA-s2 on 7 out of 28 benchmark problems. HV-based non-dominated search shows promising convergence speed in the early optimization period. However, it is prone to trap into local optima, resulting in unstable optimization performance. Classifier-assisted rank-based learning pre-screening is able to infill the non-dominated front to enhance the diversity of the non-dominated front, while local search in the sparse objective space is able to explore the uncertain but promising front. Although CLMEA-s1 and CLMEA-s3 cannot achieve promising results, the diversity of non-dominated solutions and the exploration of the uncertain but promising area are significant for the optimization even with limited computational budgets. In comparison with CLMEA-s2, CLMEA allocates more computational resources on exploring uncertain but promising area and maintaining the diversity of the solutions, which explains why CLMEA performs better than CLMEA-s2 on most benchmark functions.

### D. Experimental Results on Bi-Objective DTLZ Problems

CLMEA is compared with 5 state-of-the-art surrogate-assisted MOEAs (i.e., CPS-MOEA, K-RVEA, CSEA, END-ARMOEA, and MCEA/D) to test its performance on bi-objective DTLZ problems with dimension 30, 50, 100, and 200. Table III presents the average IGD results of 20 independent runs on bi-objective DTLZ problems, with the best results highlighted. Fig. 6 shows the convergence curves of CLMEA and compared algorithms on bi-objective 100D DTLZ problems. The convergence curves on 30D, 50D, and 200D DTLZ problems can be found in supplementary materials (Fig. S-1, S-2, and S-3). The filled color areas indicate the variance of the algorithm. Fig. 7 (a) presents the algorithm rankings for the bi-objective DTLZ problems with dimensions ranging from 30D to 200D. It can be observed that CLMEA achieves best optimization result on 25 out of 28 DTLZ problems except for 30-100D DTLZ7 problems (slightly worse than K-RVEA). The PF of DTLZ7 is discontinuous, and it is hard to maintain the diversity of non-dominated solutions. When dealing with 200D DTLZ7 problem, the performance of CLMEA becomes better than K-RVEA. MCEA/D also shows promising performance except on DTLZ7 problems. MCEA/D shows promising performance except for DTLZ7 problems, while MCEA/D is only better than CPS-MOEA on DTLZ7 problems due to the discontinuous PF. K-RVEA achieves the best results on 30D-100D DTLZ7 problems, yet performs worse than other algorithms on most of the other DTLZ7 benchmark functions.

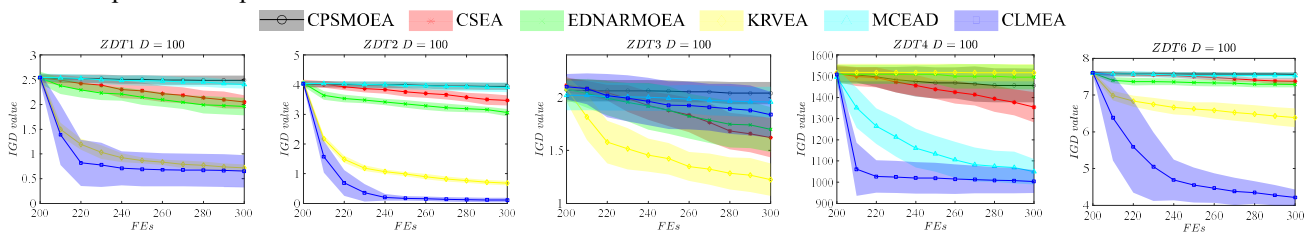

Fig. 9. Convergence curves of the six compared algorithms on 2-objective 100D ZDT problems

TABLE IV
AVERAGE IGD RESULTS OF THE 20 INDEPENDENT RUNS ON BI-OBJECTIVE ZDT PROBLEMS.

| Problems | D | CPS-MOEA | K-RVEA | CSEA | EDN-ARMOEA | MCEA/D | CLMEA |
|---|---|---|---|---|---|---|---|
| ZDT1 | 30 | 2.0868e+00 | **5.7318e-02** | 1.0050e+00 | 7.6113e-01 | 1.9462e+00 | 1.7209e-01 |
|  | 50 | 2.1891e+00 | **1.4374e-01** | 1.2874e+00 | 1.1993e+00 | 2.3084e+00 | 2.5621e-01 |
|  | 100 | 2.4903e+00 | 7.1883e-01 | 2.0480e+00 | 1.9585e+00 | 2.4053e+00 | **6.5192e-01** |
|  | 200 | 2.6073e+00 | 2.5471e+00 | 2.2855e+00 | 2.4627e+00 | 2.5974e+00 | **1.1857e+00** |
| ZDT2 | 30 | 3.2559e+00 | 7.0157e-02 | 2.1445e+00 | 1.5338e+00 | 2.9519e+00 | **9.9381e-03** |
|  | 50 | 3.7150e+00 | 1.3021e-01 | 2.7810e+00 | 2.1001e+00 | 3.4710e+00 | **1.4949e-02** |
|  | 100 | 3.9450e+00 | 6.7898e-01 | 3.4708e+00 | 3.0595e+00 | 3.9134e+00 | **1.1057e-01** |
|  | 200 | 4.2395e+00 | 1.6773e+00 | 3.8334e+00 | 3.6968e+00 | 4.1575e+00 | **7.7927e-01** |
| ZDT3 | 30 | 1.7755e+00 | **1.0618e-01** | 7.9850e-01 | 7.9003e-01 | 1.5921e+00 | 8.0636e-01 |
|  | 50 | 1.8901e+00 | **5.3108e-01** | 1.0192e+00 | 1.1492e+00 | 1.9135e+00 | 1.2693e+00 |
|  | 100 | 2.0403e+00 | **1.2244e+00** | 1.6208e+00 | 1.6982e+00 | 1.9570e+00 | 1.8374e+00 |
|  | 200 | 2.1249e+00 | 2.1718e+00 | **1.8599e+00** | 2.0865e+00 | 2.1300e+00 | 2.0650e+00 |
| ZDT4 | 30 | 3.5714e+02 | 3.0280e+02 | 2.7142e+02 | 3.2959e+02 | **2.2205e+02** | 2.6938e+02 |
|  | 50 | 6.3994e+02 | 6.7729e+02 | 5.3573e+02 | 6.6042e+02 | **4.1766e+02** | 4.5928e+02 |
|  | 100 | 1.4564e+03 | 1.5170e+03 | 1.3538e+03 | 1.4961e+03 | 1.0495e+03 | **1.0032e+03** |
|  | 200 | 2.9975e+03 | 3.2350e+03 | 2.9714e+03 | 3.1888e+03 | 2.1733e+03 | **2.0531e+03** |
| ZDT6 | 30 | 7.2138e+00 | 3.5372e+00 | 6.4063e+00 | 6.1819e+00 | 7.0424e+00 | **1.6945e+00** |
|  | 50 | 7.3874e+00 | 4.8723e+00 | 6.8988e+00 | 6.7780e+00 | 7.3941e+00 | **2.5857e+00** |
|  | 100 | 7.5674e+00 | 6.3846e+00 | 7.3704e+00 | 7.2903e+00 | 7.5345e+00 | **4.2129e+00** |
|  | 200 | 7.6615e+00 | 6.9280e+00 | 7.5123e+00 | 7.4841e+00 | 7.6840e+00 | **5.6334e+00** |
| +/−/≈ |  | 0/ 20 / 0 | 5/ 15 / 0 | 1/ 19 / 0 | 0/ 20 / 0 | 2/ 18 / 0 | NA |

For a better illustration, non-dominated front by CLMEA and compared algorithms on DTLZ2 and DTLZ5 problems are shown in Fig. 8. It can be obviously observed that CLMEA outperforms the other surrogate-assisted MOEAs. Note that CLMEA captures the whole PF of 30D DTLZ2 and DTLZ5 problems. As the dimension increases, the non-dominated front becomes farther away from PF. The performance of MCEA/D is also promising in comparison with CPS-MOEA, K-RVEA, CSEA and EDN-ARMOEA, although not converged due to the limited computational budget.

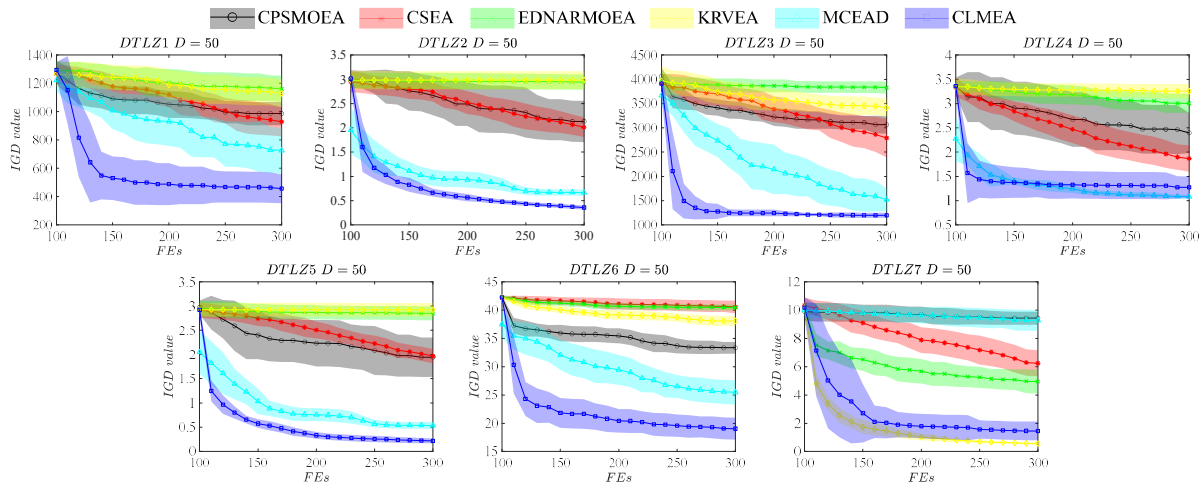

Fig. 10. Convergence curves of CLMEA and compared algorithms on 3-objective 50D DTLZ problems

TABLE V
AVERAGE IGD RESULTS OF THE 20 INDEPENDENT RUNS ON 3-OBJECTIVE DTLZ PROBLEMS.

| Problems | D | CPS-MOEA | K-RVEA | CSEA | EDN-ARMOEA | MCEA/D | CLMEA |
|---|---|---|---|---|---|---|---|
| DTLZ1 | 30 | 5.1718e+02 | 5.7816e+02 | 4.6671e+02 | 5.9632e+02 | 3.8953e+02 | **2.8640e+02** |
|  | 50 | 9.8634e+02 | 1.1282e+03 | 9.2914e+02 | 1.1605e+03 | 7.2570e+02 | **4.5536e+02** |
|  | 100 | 2.3507e+03 | 2.6747e+03 | 2.5375e+03 | 2.6373e+03 | 2.2266e+03 | **9.3303e+02** |
|  | 200 | 4.9043e+03 | 5.7369e+03 | 5.4886e+03 | 5.6628e+03 | 4.2725e+03 | **2.1620e+03** |
| DTLZ2 | 30 | 1.3452e+00 | 1.5993e+00 | 9.1154e-01 | 1.5621e+00 | 5.0009e-01 | **2.9013e-01** |
|  | 50 | 2.1332e+00 | 2.9820e+00 | 2.0117e+00 | 2.9442e+00 | 6.6442e-01 | **8.9291e-01** |
|  | 100 | 6.0256e+00 | 6.4428e+00 | 5.7078e+00 | 6.4135e+00 | 2.6861e+00 | **1.7226e+00** |
|  | 200 | 1.2476e+01 | 1.3927e+01 | 1.3169e+01 | 1.3929e+01 | 4.7093e+00 | **3.1795e+00** |
| DTLZ3 | 30 | 1.5398e+03 | 1.7148e+03 | 1.3786e+03 | 1.9712e+03 | 7.8677e+02 | **6.7609e+02** |
|  | 50 | 3.0635e+03 | 3.4132e+03 | 2.7900e+03 | 3.8282e+03 | 1.5202e+03 | **1.1955e+03** |
|  | 100 | 7.5129e+03 | 8.3472e+03 | 7.9512e+03 | 8.4993e+03 | 5.3212e+03 | **2.6658e+03** |
|  | 200 | 1.5834e+04 | 1.8563e+04 | 1.7843e+04 | 1.8479e+04 | 1.1642e+04 | **5.7911e+03** |
| DTLZ4 | 30 | 1.6353e+00 | 1.7399e+00 | 1.0017e+00 | 1.5396e+00 | 1.0014e+00 | 1.0051e+00 |
|  | 50 | 2.3978e+00 | 3.2546e+00 | 1.8663e+00 | 3.0066e+00 | 1.0832e+00 | 1.2718e+00 |
|  | 100 | 5.7630e+00 | 6.6753e+00 | 5.4100e+00 | 6.6012e+00 | 2.6458e+00 | **1.8990e+00** |
|  | 200 | 1.2024e+01 | 1.4116e+01 | 1.2442e+01 | 1.4000e+01 | 4.6499e+00 | **3.2875e+00** |
| DTLZ5 | 30 | 1.1801e+00 | 1.4422e+00 | 9.4208e-01 | 1.4853e+00 | 3.2371e-01 | **1.8035e-01** |
|  | 50 | 1.9458e+00 | 2.9230e+00 | 1.9672e+00 | 2.8475e+00 | 5.3391e-01 | **2.1626e-01** |
|  | 100 | 5.2995e+00 | 6.3502e+00 | 5.6636e+00 | 6.2066e+00 | 2.3666e+00 | **1.4172e+00** |
|  | 200 | 1.0529e+01 | 1.3788e+01 | 1.3233e+01 | 1.3832e+01 | 4.8062e+00 | **3.2023e+00** |
| DTLZ6 | 30 | 1.8583e+01 | 1.9859e+01 | 2.2652e+01 | 2.2927e+01 | 1.3639e+01 | **9.3408e+00** |
|  | 50 | 3.3332e+01 | 3.8063e+01 | 4.0560e+01 | 4.0444e+01 | 2.5471e+01 | **1.9042e+01** |
|  | 100 | 7.4714e+01 | 8.3046e+01 | 8.5935e+01 | 8.5176e+01 | 6.4448e+01 | **4.4918e+01** |
|  | 200 | 1.5628e+02 | 1.7365e+02 | 1.7571e+02 | 1.7422e+02 | 1.3504e+02 | **1.0259e+02** |
| DTLZ7 | 30 | 8.6688e+00 | **2.4665e-01** | 5.2500e+00 | 3.7350e+00 | 7.7191e+00 | 8.0399e-01 |
|  | 50 | 9.4286e+00 | **5.6056e-01** | 6.2350e+00 | 4.9518e+00 | 9.2709e+00 | 1.4570e+00 |
|  | 100 | 1.0523e+01 | **2.8565e+00** | 8.8200e+00 | 7.0809e+00 | 1.0136e+01 | 3.5307e+00 |
|  | 200 | 1.0853e+01 | 1.0801e+01 | 9.6934e+00 | 9.2666e+00 | 1.0791e+01 | **6.3128e+00** |
| $+/-/\approx$ |  | 0/ 28 / 0 | 3/ 25 / 0 | 0/ 27 / 1 | 0/ 28 / 0 | 1/ 26 / 1 | NA |

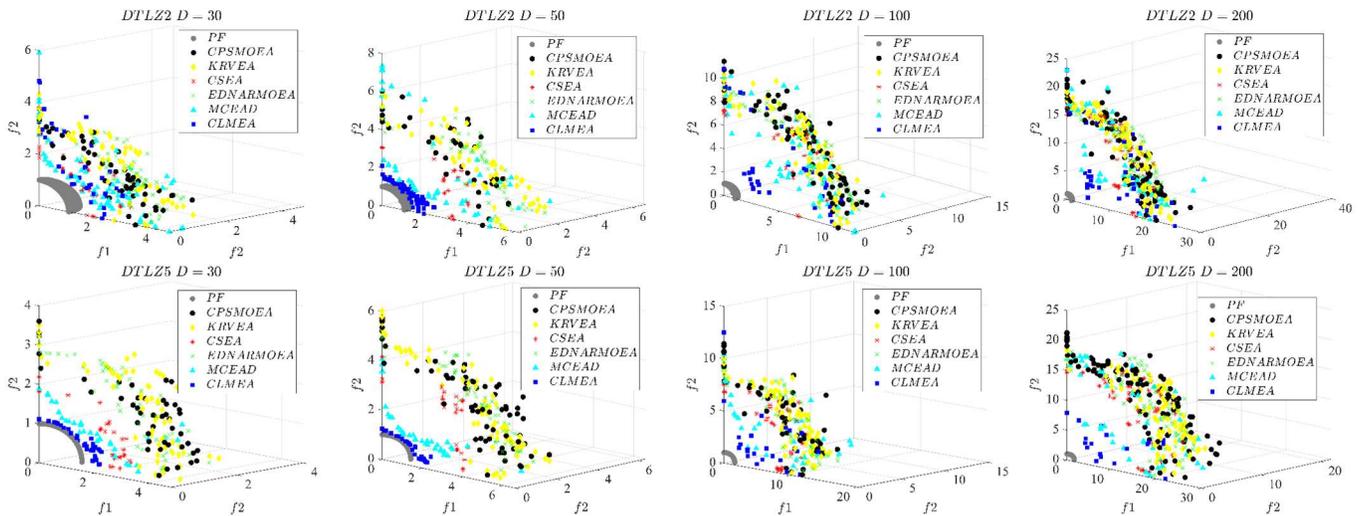

Fig. 11. Non-dominated front by CLMEA and compared algorithms on 3-objective DTLZ problems

*E. Experimental Results on Bi-Objective ZDT Problems*

ZDT is then employed to further test the performance of CLMEA and the compared algorithms. Table IV shows the average IGD results of 20 independent runs on bi-objective ZDT problems, with the best results highlighted. Fig. 9 presents the convergence curves of CLMEA and compared algorithms on bi-objective 100D DTLZ problems. The convergence curves on 30D, 50D, and 200D ZDT problems can be found in supplementary materials (Fig. S-4, S-5, and S-6). CLMEA performs better than other algorithms on most benchmark functions except ZDT3, since ZDT3 is a multimodal problem and the PF is discrete. The performance of CLMEA on high-dimensional ZDT1 and ZDT4 is better, in comparison with other algorithms. Despite poor performance on ZDT4, K-RVEA also shows promising performance on ZDT problems. To take a further look at the final solution distributions, Fig. 8 illustrates the final non-dominated front of ZDT2 from 30D to 200D. CLMEA captures the entire PF on 30D ZDT2 problem. The solutions of CLMEA on PF become sparse when the dimension increases to 50D and 100D. When the dimension increases to 200, the non-dominated solutions of CLMEA become far from the PF. Fig. 7 (b) presents the violin plot of algorithm rankings for the bi-objective DTLZ problems with dimensions ranging from 30D to 200D. From the algorithm rankings, CLMEA achieves the best performance on most ZDT problems. CPS-MOEA and MCEA/D obtain worse performance than the other algorithms on ZDT problems.

*F. Experimental Results on 3-Objective DTLZ Problems*

Results on 3-objective ZDT problems are present in Table V. Fig. 10 illustrates convergence curves of CLMEA and compared algorithms on 3-objective 50D ZDT problems (The convergence curves on 30D, 100D, and 200D DTLZ problems can be found in Fig. S-7, S-8, and S-9 of the supplementary materials). CLMEA achieves the best average IGD value on 23 out of 28 problems in comparison with other 5 algorithms. Since the PF of DTLZ7 is discontinuous, and it is hard to maintain the diversity of non-dominated solutions, resulting in CLMEA worse than K-RVEA. K-RVEA only shows best performance on 30D, 50D, and 100D DTLZ7 problems.

Fig. 11 shows the final solution distributions of the algorithms on 3-objective DTLZ problems. For problems from 30 to 50D, CLMEA provides non-dominated solutions close to PF with great diversity. When the dimension increases to 100 and 200D, the non-dominated solutions become relatively sparse and far to PF. MCEA/D also presents promising non-dominated front that is closer to PF than other compared algorithms except CLMEA. The violin plot for the algorithm rankings is shown in Fig. 7 (c). It can be observed that K-RVEA and EDN-ARMOEA present worst algorithm rankings, while CLMEA and MCEAD show promising search abilities.

*G. Computational Complexity Analysis*

Computational time for training surrogates and selecting infill sample candidates varies with surrogate-assisted methods. The entire time for the optimization process includes the computational time on training the surrogate, selecting the infill candidates, and conducting real FEs. The computational complexity for training RBF and PNN is on the order of $O(mn^2D)$, and for each prediction is $O(mnD)$, where $m$ is the number of objectives, $n$ is the number of training samples, and $D$ is the dimension of the optimization problem. For training Kriging model, the computational complexity is $O(mn^3D)$ [30], which is time consuming and sensitive to the number of training sample points.

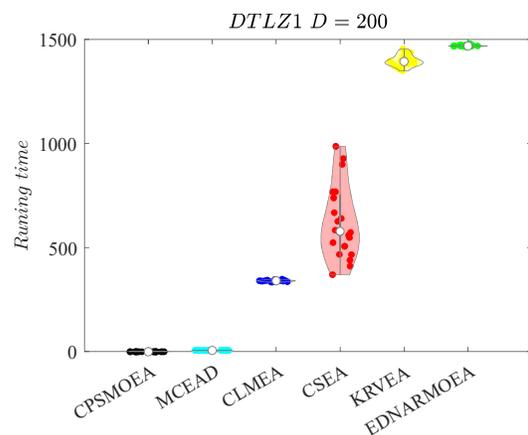

Fig. 12. Runtime of CLMEA and compared algorithms for 300 FEs on bi-objective 200D ZDT1 problems

To investigate the computational complexity of CLMEA and compared algorithms, computational times of all the six algorithms on bi-objective 200D DTLZ1 with 300 FEs are compared in Fig. 12. The computational times for CPS-MOEA and MCEA/D are negligible, which are 0.05s and 6.48s, respectively. CLMEA takes slightly less time to run than CSEA. K-RVEA and END-ARMOEA consume the most runtime due to the costly training of the Kriging and dropout neural network. Since the real-world simulation time is hours even days, the computational time for training surrogates and selecting infill samples can be negligible. When each FE is not that expensive, MCEA/D is a good choice due to the promising optimization ability and low computational complexity.

*H. Application on Geothermal Development System*

One of the potential application scenarios of the proposed CLMEA algorithm is expected to be the field of the complex engineering design that involving a series of objectives and numerical simulations. As a reliable renewable and sustainable energy, geothermal resource development plays a significant role in the transition from fossil fuels. A heat extraction optimization problem of enhanced geothermal system (EGS) is employed to further test the efficacy of the proposed algorithm. Geothermal resources development aims to maximize the heat extraction by injecting cold water and producing heated water. The decision variables to be optimized are the water-injection rates and water-production rates (or bottom hole pressure) of wells. Maximizing only the long-term revenue may takes risks due to geological uncertainties. To reduce the risks in the reservoir management, short-term potential should also be considered. For a given heat extraction optimization problem, the goal is to maximize the long- and short-term life-cycle net present value:

$$\max_{\boldsymbol{x}} f_l(\boldsymbol{x}, \boldsymbol{z}_l) \qquad (14)$$
$$\max_{\boldsymbol{x}} f_s(\boldsymbol{x}, \boldsymbol{z}_s)$$

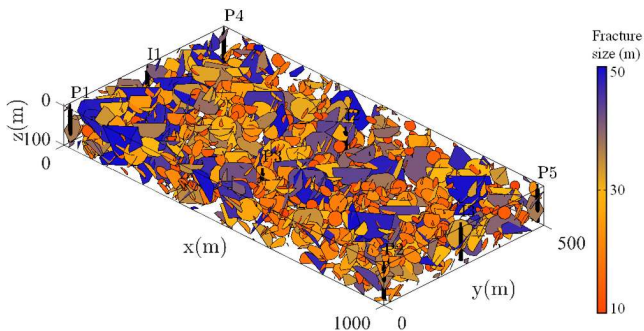

Fig. 13. Discrete fracture network and well-placement distribution of the field-scale EGS.

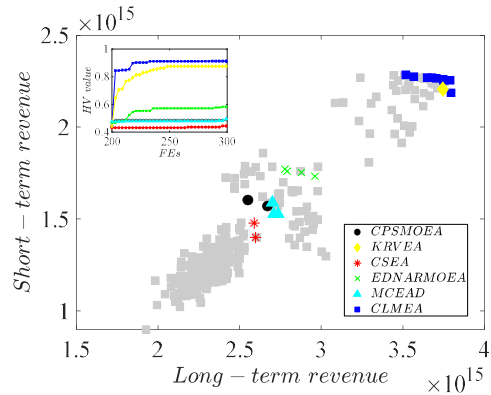

Fig. 14. Non-dominated solutions and corresponding HV values obtained by different algorithms on real-world heat extraction optimization of EGS.

s.t.
$$\boldsymbol{lb} \leqslant \boldsymbol{x} \leqslant \boldsymbol{ub},\ \boldsymbol{x} \in \mathbb{R}^d \qquad (15)$$

$$f(\boldsymbol{x}, \boldsymbol{z}) = r_e CTEP(\boldsymbol{x}, \boldsymbol{z}) - r_i CWI(\boldsymbol{x}, \boldsymbol{z}) - r_p CWP(\boldsymbol{x}, \boldsymbol{z}) \qquad (16)$$

where $\boldsymbol{x}$ is the decision vector to be optimized, $\boldsymbol{z}$ is the state vector (i.e., the temperature and pressure of the fractures at each grid block) with subscript $l$ representing long-term and $s$ representing short-term, $r_e$, $r_i$ and $r_p$ represent price of thermal energy, the water-injection cost and water-production cost respectively, and $CTEP$, $CWI$ and $CWP$ denote the cumulative thermal energy production, cumulative water injection, and cumulative water production respectively. More details about the problem can be found in [64, 65].

In this case, the geological model contains large number of fractures, and the simulation for corresponding numerical model is time-consuming. The discrete fracture network and well-placement distribution of the field-scale EGS is present in Fig. 13. This problem contains three injection wells and five production wells, as indicated in Fig. 13. The life time of the project is 6000 days, and the time-step length is set to 300 days. Thus, the problem totally involves 160 decision variables and two objective functions.

Fig. 14 shows non-dominated solutions and corresponding HV values obtained by different algorithms on real-world heat extraction optimization of EGS. The gray square points are the evaluated points of CLMEA which can reveal the convergence and exploration property of the algorithm. Notably, CLMEA achieves best non-dominated solutions and holds best HV value after 300 simulation evaluations. K-RVEA also shows promising HV value. Nevertheless, the diversity of K-RVEA is low, with only one final non-dominated solution. For MCEA/D, since the ideal point for real-world case is unknown, it is hard to distribute the weighted vector, resulting in poor performance on this case.

V. CONCLUSION

In this paper, a classifier-assisted rank-based learning and local model based multi-objective evolutionary algorithm, called CLMEA, has been proposed to solve high-dimensional expensive multi-objective optimization problems. CLMEA contains three sampling strategies: classifier-assisted rank-based learning pre-screening, HV-based non-dominated search, and local search in the sparse objective space. classifier-assisted rank-based learning pre-screening strategy adopts a PNN as classifier to divide the offspring into a number of ranks. The offspring uses rank-based learning strategy to generate more promising and informative candidates for real FEs. HV-based non-dominated search strategy employs RBF as surrogate. After searching Pareto solutions of surrogate model with evolutionary algorithm, the candidates with higher HV improvement are selected for real FEs. To keep the diversity of solutions, the sparse sample points from the Pareto solutions is selected as the guided parent and center of the local surrogate to further infill in the uncertain region of the Pareto front.

CLMEA was compared with five state-of-the-art algorithms, i.e., CPSMOEA, KRVEA, END-ARMOEA, CSEA, and MCEA/D. The results show its superiority on various test suites, i.e., DTLZ and ZDT benchmark problems, in comparison with state-of-the-art algorithms. In addition, the proposed algorithm also shows promising results in practical application related to geothermal reservoir heat extraction optimization.

In our future work, learning-aided offspring generation will be studied, aiming to construct an effective actor to generate elite offspring and extend to single-objective and multi-modal problems. Since many real-world applications and engineering design involve computationally expensive simulation, work will primarily focus on optimization with limited computational budgets.

# Supplementary material for "Rank-Based Learning and Local Model Based Evolutionary Algorithm for High-Dimensional Expensive Multi-Objective Problems"

Guodong Chen, Jiu Jimmy Jiao, Xiaoming Xue and Zhongzheng Wang

TABLE S-I
CHARACTERISTICS OF THE DTLZ AND ZDT BENCHMARK PROBLEMS.

| Problem | M | D | Separability | Modality | Bias | Geometry |
|---|---|---|---|---|---|---|
| DTLZ1 | 2, 3 | 30, 50, 100, 200 | Separable | M | No | Linear |
| DTLZ2 | 2, 3 | 30, 50, 100, 200 | Non-separable | U | No | Concave |
| DTLZ3 | 2, 3 | 30, 50, 100, 200 | Separable | M | No | Concave |
| DTLZ4 | 2, 3 | 30, 50, 100, 200 | Separable | U | Yes | Concave, many-to-one |
| DTLZ5 | 2, 3 | 30, 50, 100, 200 | Separable | U | Yes | Concave |
| DTLZ6 | 2, 3 | 30, 50, 100, 200 | Non-separable | U | Yes | Concave |
| DTLZ7 | 2, 3 | 30, 50, 100, 200 | Non-separable | M | Yes | Disconnected |
| ZDT1 | 2 | 30, 50, 100, 200 | Separable | U | No | Convex |
| ZDT2 | 2 | 30, 50, 100, 200 | Separable | U | No | Concave |
| ZDT3 | 2 | 30, 50, 100, 200 | Separable | M | No | Convex, disconnected |
| ZDT4 | 2 | 30, 50, 100, 200 | Separable | M | No | Convex, multi-modal |
| ZDT6 | 2 | 30, 50, 100, 200 | Separable | M | Yes | Convex, multi-modal, many-to-one |

TABLE S-II
AVERAGE IGD VALUES OF CLMEA AND ITS VARIANTS ON BI-OBJECTIVE DTLZ PROBLEMS.

| Problems | D | CLMEA-s1 | CLMEA-s2 | CLMEA-s3 | CLMEA |
|---|---|---|---|---|---|
| DTLZ1 | 30 | 4.7253e+02 | 3.8505e+02 | 5.4359e+02 | **3.0310e+02** |
| | 50 | 8.9219e+02 | 5.8584e+02 | 1.0991e+03 | **5.3412e+02** |
| | 100 | 2.4696e+03 | 1.3976e+03 | 2.5042e+03 | **1.0740e+03** |
| | 200 | 5.2544e+03 | 2.8283e+03 | 4.9325e+03 | **1.9906e+03** |
| DTLZ2 | 30 | 1.1418e+00 | 1.0045e-01 | 4.9472e-01 | **9.7035e-02** |
| | 50 | 2.0315e+00 | 1.6594e-01 | 1.0468e+00 | **1.3942e-01** |
| | 100 | 4.5559e+00 | 4.2610e-01 | 3.2457e+00 | **3.8858e-01** |
| | 200 | 9.3829e+00 | 1.2610e+00 | 7.1683e+00 | **1.1857e+00** |
| DTLZ3 | 30 | 1.1057e+03 | 7.2571e+02 | 1.3641e+03 | **7.0538e+02** |
| | 50 | 2.0866e+03 | 1.2179e+03 | 2.8047e+03 | **1.1711e+03** |
| | 100 | 6.2104e+03 | 3.1289e+03 | 6.0766e+03 | **2.6277e+03** |
| | 200 | 1.3579e+04 | **5.2042e+03** | 1.1243e+04 | 5.2284e+03 |
| DTLZ4 | 30 | 1.3643e+00 | 7.0425e-01 | 8.8482e-01 | **6.7771e-01** |
| | 50 | 2.4089e+00 | **7.2529e-01** | 1.3839e+00 | 7.3285e-01 |
| | 100 | 4.8569e+00 | **7.8942e-01** | 3.5566e+00 | 8.8932e-01 |
| | 200 | 9.5625e+00 | 1.0446e+00 | 7.3095e+00 | **9.5823e-01** |
| DTLZ5 | 30 | 1.1151e+00 | 1.7222e-01 | 5.4208e-01 | **1.0114e-01** |
| | 50 | 2.0741e+00 | 1.9707e-01 | 1.0306e+00 | **1.4158e-01** |
| | 100 | 4.6873e+00 | 5.6684e-01 | 3.3421e+00 | **4.6926e-01** |
| | 200 | 9.7173e+00 | 2.2630e+00 | 7.0341e+00 | **1.3569e+00** |
| DTLZ6 | 30 | 1.3817e+01 | **9.3975e+00** | 2.2244e+01 | 1.1261e+01 |
| | 50 | 2.6529e+01 | **1.8215e+01** | 3.8546e+01 | 2.0184e+01 |
| | 100 | 6.6583e+01 | **4.4500e+01** | 7.8716e+01 | 4.6223e+01 |
| | 200 | 1.4102e+02 | **9.2870e+01** | 1.5782e+02 | 9.5831e+01 |
| DTLZ7 | 30 | 3.5438e+00 | 4.9388e-01 | 4.8013e+00 | **4.7586e-01** |
| | 50 | 4.3019e+00 | 9.0252e-01 | 5.6669e+00 | **8.9416e-01** |
| | 100 | 6.4012e+00 | 2.4957e+00 | 6.3810e+00 | **1.8606e+00** |
| | 200 | 6.8865e+00 | 5.7150e+00 | 7.0836e+00 | **5.0233e+00** |
| $+/-/\approx$ | | 0/ 28/ 0 | 6/ 21/ 1 | 0/ 28/ 0 | NA |

TABLE S-III
AVERAGE IGD VALUES OF CLMEA AND ITS VARIANTS ON BI-OBJECTIVE ZDT PROBLEMS.

| Problems | D | CLMEA-s1 | CLMEA-s2 | CLMEA-s3 | CLMEA |
|---|---|---|---|---|---|
| ZDT1 | 30 | 1.1319e+00 | 1.9790e-01 | 1.6613e+00 | **1.7209e-01** |
|  | 50 | 1.5609e+00 | 1.9546e-01 | 1.8696e+00 | **2.5621e-01** |
|  | 100 | 2.2031e+00 | 3.4978e-01 | 2.2699e+00 | **6.5192e-01** |
|  | 200 | 2.4274e+00 | 1.2138e+00 | 2.4514e+00 | **1.1857e+00** |
| ZDT2 | 30 | 2.0358e+00 | 2.3865e-02 | 2.8708e+00 | **9.9381e-03** |
|  | 50 | 2.5258e+00 | 6.4437e-02 | 3.3187e+00 | **1.4949e-02** |
|  | 100 | 3.5770e+00 | 1.3884e-01 | 3.8033e+00 | **1.1057e-01** |
|  | 200 | 3.9506e+00 | 1.1914e+00 | 4.0711e+00 | **7.7927e-01** |
| ZDT3 | 30 | 1.1143e+00 | 8.8016e-01 | 1.5004e+00 | **8.0636e-01** |
|  | 50 | 1.3314e+00 | 1.1862e+00 | 1.6275e+00 | **1.2693e+00** |
|  | 100 | 1.9137e+00 | 1.7946e+00 | 1.9108e+00 | **1.8374e+00** |
|  | 200 | 2.0254e+00 | 1.8959e+00 | 2.0386e+00 | **2.0650e+00** |
| ZDT4 | 30 | 3.3374e+02 | 2.8070e+02 | 2.7557e+02 | **2.6938e+02** |
|  | 50 | 6.2420e+02 | 4.6421e+02 | 5.4145e+02 | **4.5928e+02** |
|  | 100 | 1.3618e+03 | **9.4616e+02** | 1.3254e+03 | 1.0032e+03 |
|  | 200 | 2.8201e+03 | **2.0295e+03** | 2.8041e+03 | 2.0131e+03 |
| ZDT6 | 30 | 6.3961e+00 | **1.5703e+00** | 6.9063e+00 | 1.6945e+00 |
|  | 50 | 6.8971e+00 | **2.1674e+00** | 7.1745e+00 | 2.5857e+00 |
|  | 100 | 7.4228e+00 | 5.6708e+00 | 7.4932e+00 | **4.2129e+00** |
|  | 200 | 7.5752e+00 | 6.2151e+00 | 7.5844e+00 | **5.6334e+00** |
| $+/-/\approx$ |  | 0 / 20 / 0 | 3 / 16 / 1 | 0 / 20 / 0 | NA |

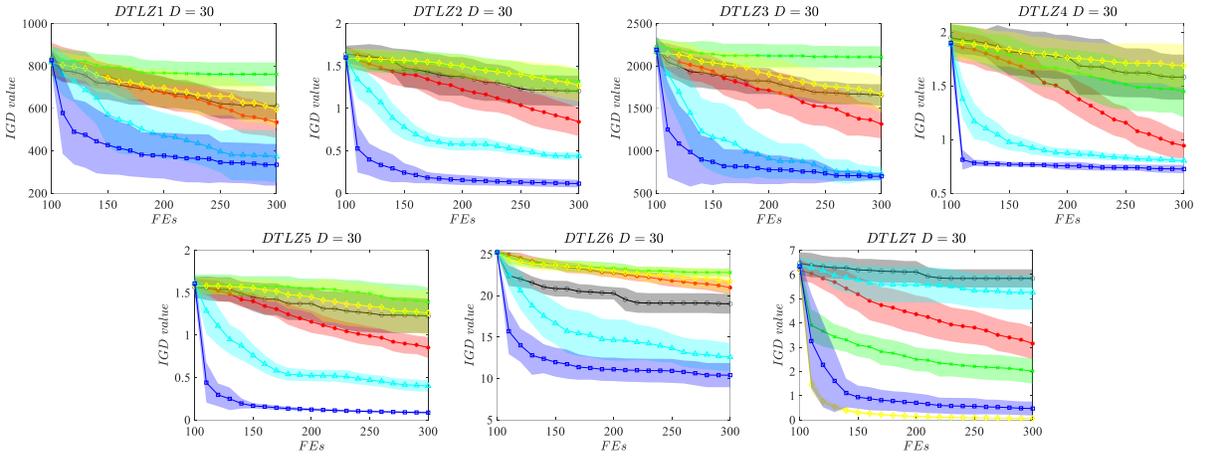

Fig. S-1. Convergence curves of the six compared algorithms on 2-objective 30D DTLZ problems

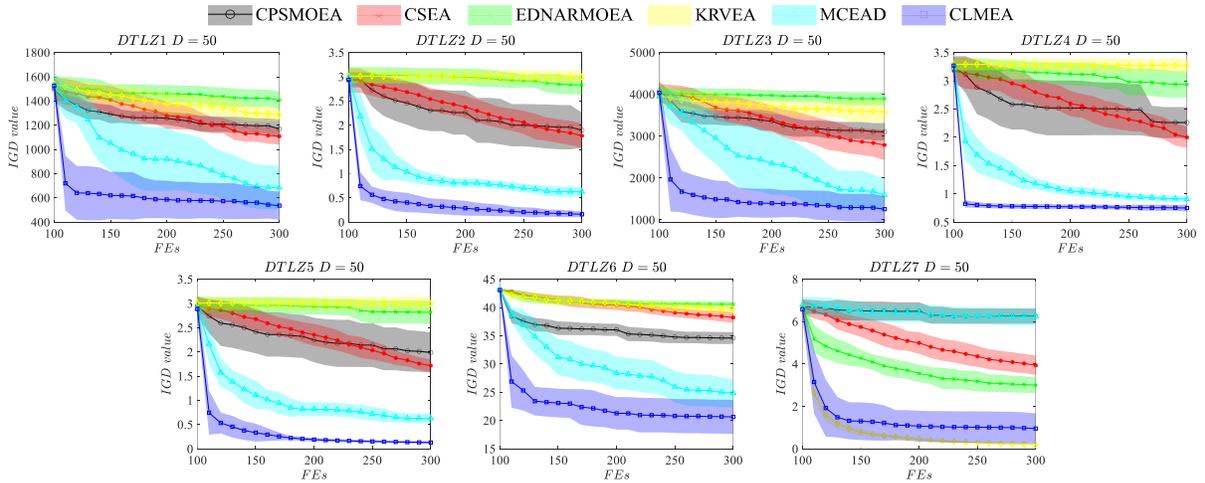

Fig. S-2. Convergence curves of the six compared algorithms on 2-objective 50D DTLZ problems

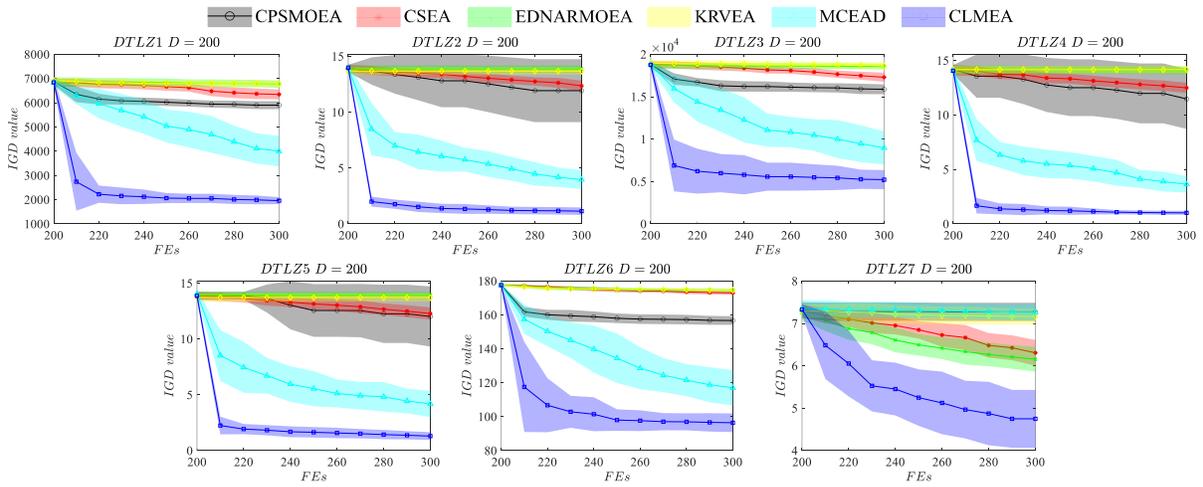

Fig. S-3. Convergence curves of the six compared algorithms on 2-objective 200D DTLZ problems

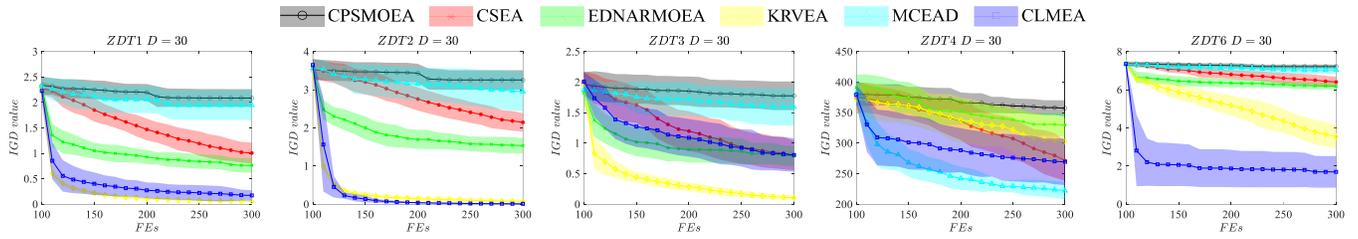

Fig. S-4. Convergence curves of the six compared algorithms on 2-objective 30D ZDT problems

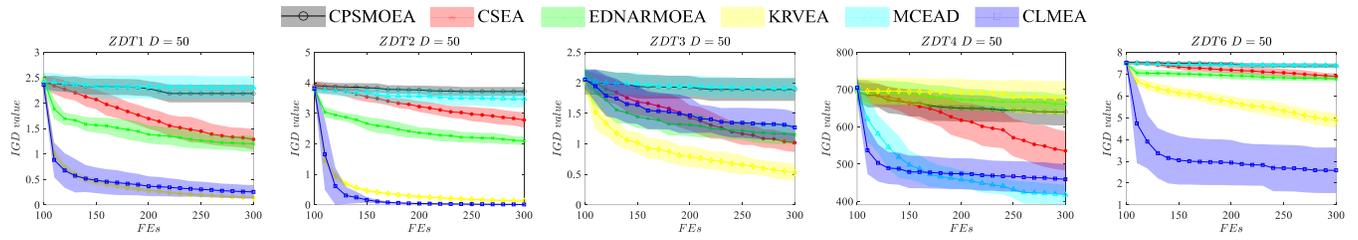

Fig. S-5. Convergence curves of the six compared algorithms on 2-objective 50D ZDT problems

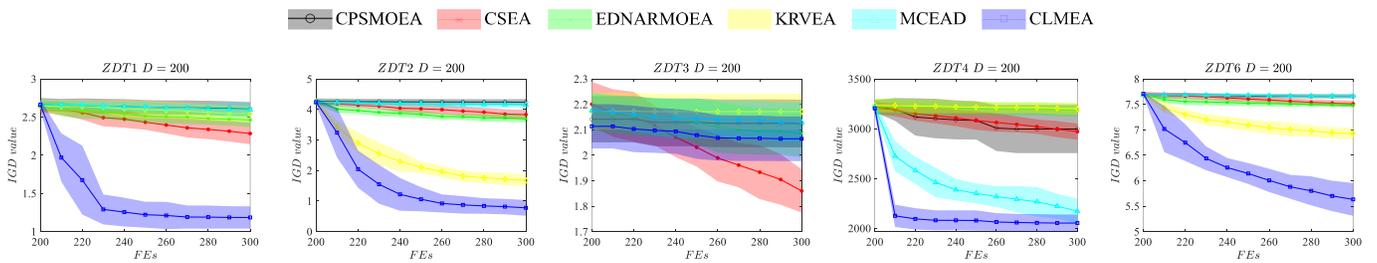

Fig. S-6. Convergence curves of the six compared algorithms on 2-objective 200D ZDT problems

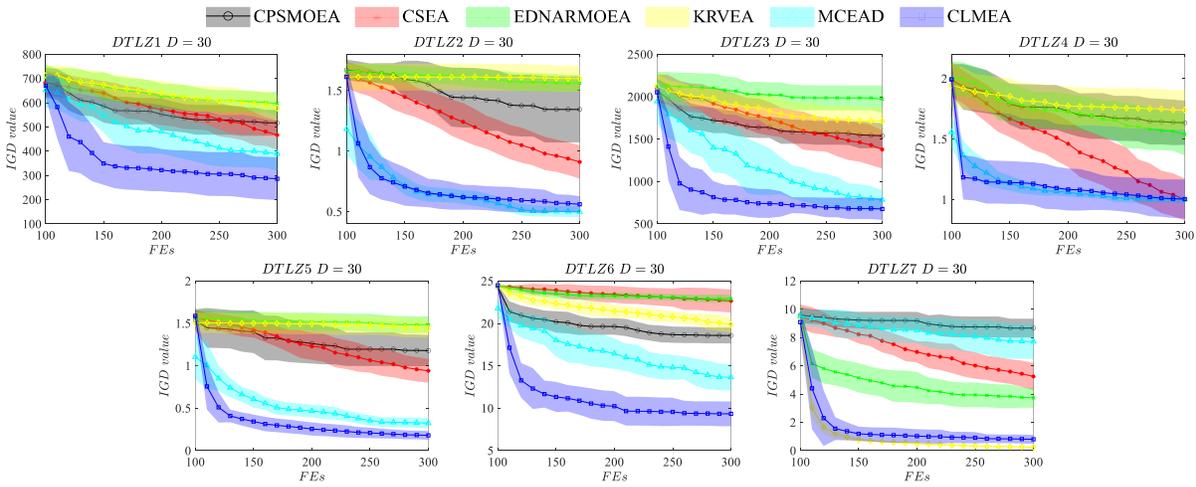

Fig. S-7. Convergence curves of the six compared algorithms on 3-objective 30D DTLZ problems

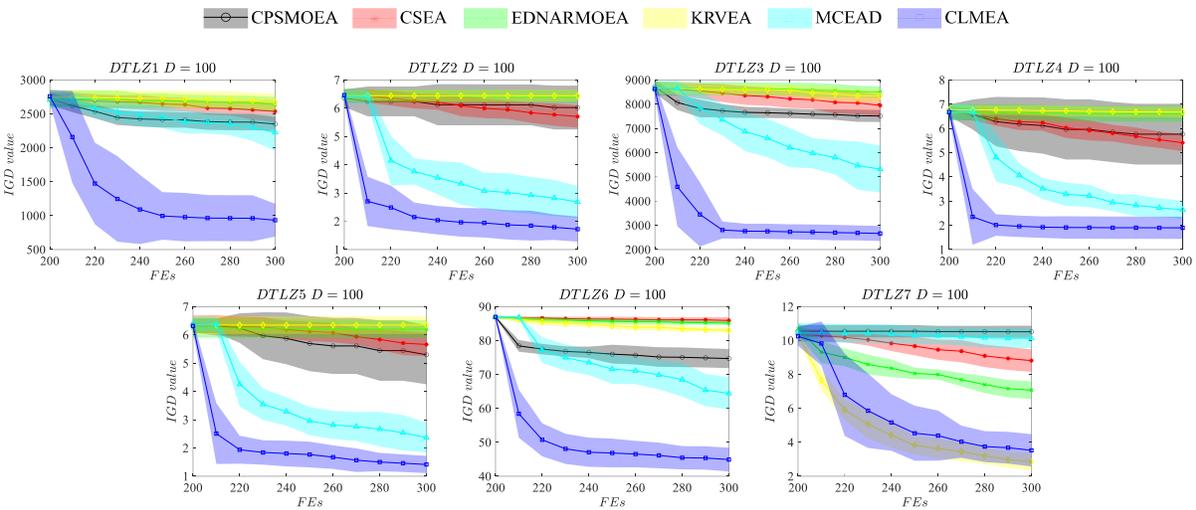

Fig. S-8. Convergence curves of the six compared algorithms on 3-objective 100D DTLZ problems

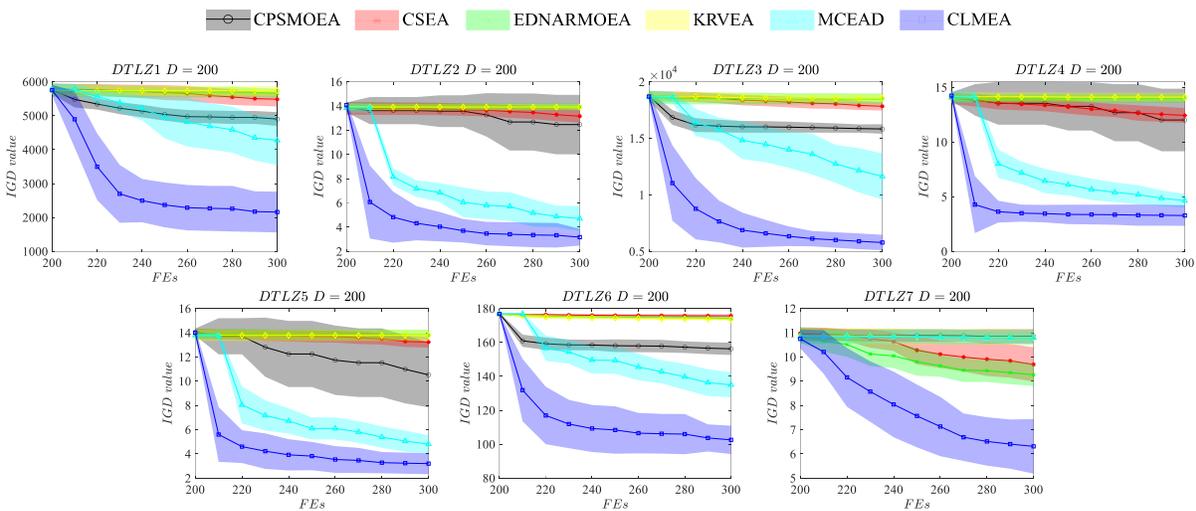

Fig. S-9. Convergence curves of the six compared algorithms on 3-objective 200D DTLZ problems